\PassOptionsToPackage{noend}{algorithmic}
\documentclass[journal,12pt,onecolumn,draftclsnofoot,letterpaper]{IEEEtran}
\usepackage{microtype,graphicx,booktabs}
\usepackage[numbers,sort&compress]{natbib}
\usepackage{amsmath,amssymb,mathtools,amsthm}
\usepackage{algorithm,algorithmic,float,multirow,dsfont}
\usepackage{xcolor,url}
\usepackage{placeins}
\usepackage{hyperref}

\usepackage[capitalize,noabbrev]{cleveref}
\theoremstyle{plain}
\newtheorem{theorem}{Theorem}[section]
\newtheorem{proposition}[theorem]{Proposition}
\newtheorem{corollary}[theorem]{Corollary}
\theoremstyle{definition}
\newtheorem{definition}[theorem]{Definition}
\newtheorem{property}[theorem]{Property}
\newtheorem{assumption}[theorem]{Assumption}
\theoremstyle{plain}
\newtheorem{remark}[theorem]{Remark}
\usepackage{soul}

\usepackage{xcolor}
\definecolor{MediumSlateBlue}{HTML}{0000CD}
\newcommand{\dalimprove}[1]{\textcolor{MediumSlateBlue}{#1}}

\begin{document}
\title{Detection Augmented Learning In Non-Stationary Bandit Environments}
\author{Argyrios~Gerogiannis, Yu-Han~Huang, Subhonmesh~Bose,
        and~Venugopal~V.~Veeravalli%
\thanks{The authors are with the ECE department and the Coordinated Science Laboratory, Grainger College of Engineering, University of Illinois at Urbana-Champaign. Email:\{ag91,yuhanhh2,boses,vvv\}@illinois.edu. An earlier version of this work appeared in
\cite{gerogiannis2026dal}. This work  was supported in part by a grant from the C3.ai Digital Transformation Institute, and in part by the Army Research Laboratory under Cooperative Agreement W911NF-17-2-0196, through the University of Illinois at Urbana-Champaign.}

}
%\markboth{Preprint}%
%{Gerogiannis \MakeLowercase{\textit{et al.}}: Detection Augmented Learning}
\maketitle
\begin{abstract}
We study the problem of non-stationary bandits without knowledge of the environmental changes. We propose Detection Augmented Learning (DAL) a black-box paradigm that comprises three modular components: a bandit algorithm, a change detector, and a forced exploration schedule. This modular design enables its applicability across a broad spectrum of bandit problems, from classical to highly structured settings. First, we provide theoretical insights into DAL's strong empirical performance: a single regret bound shows that DAL transfers the regret of its stationary input to the piecewise stationary setting, attaining or improving the best known bounds in every setting considered. We further identify conditions under which DAL performs well in drifting environments. Then, we showcase DAL's practical performance and versatility through extensive experimentation in which DAL consistently surpasses all state-of-the-art methods, both prior-free and prior-based, across diverse non-stationary scenarios in nine bandit settings. Finally, we investigate the effectiveness of the forced exploration and show that a bandit-driven strategy improves performance and shortens detection delay.
\end{abstract}

\section{Introduction}
\label{sec:intro}

\IEEEPARstart{B}{andit} models underpin a wide range of engineering systems, from recommendation to real-time bidding~\citep{li2010recomm,flajolet2017bidding}. Many variants of bandits have emerged since the work of \cite{robbins1952mab}, including parametric bandits (PBs)~\citep{auer2002linearb,faury2020concord,filipi2010genlb}, non-parametric bandits~\citep{srinivas210KernBandits,bubeck2011xarmed}, contextual bandits (CBs)~\citep{woodroofe1976contextual,langford2007contextual}, combinatorial bandits~\citep{chen2013cmab,kveton2015tight}, and bandits with graph feedback~\citep{caron2012side}. In the general bandit problem, in each round, an agent receives a context $C_t$ randomly sampled from a set $\mathcal{C}$ according to $\mathcal P_t$, and chooses action $A_t=\pi_t(C_t)$ from a compact action set $\mathcal{A} \subseteq \mathbb{R}^d$ using a causal policy $\pi_t$. We compare the agent with a policy set $\Pi$---a set of mappings from $\mathcal{C}$ to $\mathcal A$. The agent receives reward
$
 X_t=f_t(C_t,A_t)+\varepsilon_t,
$
where $f_t:\mathcal{C}\times\mathcal{A}\rightarrow\mathbb{R}$ is the mean reward function and $\varepsilon_t$ is conditionally zero-mean sub-Gaussian noise. Depending on the feedback model, the agent may also observe individual item outcomes or the rewards of other actions, likewise with conditionally sub-Gaussian noise. The goal is to minimize the dynamic regret, using the policy $\pi_t$:
\begin{equation}
 R_T:=\mathbb E\!\left[\sum_{t=1}^T
 \max_{\pi\in\Pi}f_t(C_t,\pi(C_t))-f_t(C_t,A_t)\right].
 \label{eq:dynamic_regret}
\end{equation}

CBs follow the general formulation above, where the context $C_t$ is independently sampled from $\mathcal P_t$ and $|\mathcal A|$ is finite. In the other settings, the context is fixed across time and, with slight abuse of notation, we write $f_t(C_t,A_t)=f_t(A_t)$. For parametric bandits, $f_t(A_t)=\mu(\langle\theta_t,A_t\rangle)$, where $\theta_t\in\mathbb R^d$ is a bounded unknown parameter, and $\mu:\mathbb R\to\mathbb R$ is injective. These include linear bandits (LBs), with $\mu$ as identity, generalized linear bandits (GLBs), and self-concordant bandits (SCBs),  where \(\mu\) is self-concordant and the noise variance may depend on the mean~\citep{russac2021nsweightscbs}. 

For non-parametric bandits, we consider kernelized bandits (KBs) and H\"older bandits (HBs). In KBs $f_t\in H_k$, a reproducing kernel Hilbert space (RKHS) induced by a continuous positive semi-definite kernel $k:\mathcal A\times\mathcal A\to\mathbb R$ with $k(a,a)\le1$ and $\|f_t\|_{H_k}\le B$ and we define the maximum information gain $\gamma_T$ to be the worst-case mutual information between $f$ and $T$ noisy evaluations~\citep{srinivas210KernBandits}. In HBs, $\mathcal A=[0,1]^d$ and $|f_t(a)-f_t(a')|\le L\|a-a'\|_\infty^\beta$ with $0<\beta\le1$, where $L>0$ is the smoothness constant, and $\beta$ is the H\"older exponent; $\beta=1$ recovers Lipschitz rewards. The instance-dependent complexity is described by a common near-optimality packing exponent $d_z\ge0$~\citep{bubeck2011xarmed}. It controls the largest number of pairwise $\varepsilon$-separated, $\mathcal O(\varepsilon)$-optimal actions under the metric $\rho_\beta(a,a')=L\|a-a'\|_\infty^\beta$: this number is $\mathcal O(\varepsilon^{-d_z})$, with common constants across stationary segments. The choice $d_z=d/\beta$ is always admissible for the H\"older class, while a smaller exponent captures favorable geometry near the optimum.

In combinatorial bandits, actions form a feasible family $\mathcal A$ of subsets of at most $K$ of $d$ items, so, with slight abuse in notation, $d$ is the number of items, $K$ is the maximum action size, and $|\mathcal A|$ counts feasible subsets. Their mean reward is $f_t(a)=\sum_{j\in a}\theta_{t,j}$, with $\theta_t\in[0,1]^d$. We consider combinatorial semi-bandits (CSBs), where every selected item's outcome is observed, and combinatorial full-bandits (CFBs), where only the sum of these outcomes is observed. Both have linear mean rewards on the incidence vectors $x_a=(\mathds1\{j\in a\})_{j=1}^d$, but differ in the feedback available for learning and detection. For CFBs, let $r=\dim\operatorname{span}\{x_a:a\in\mathcal A\}\le d$
denote the incidence-span dimension and we consider conditionally
independent item noises with uniformly bounded sub-Gaussian
variance, so the observed sum has a common variance $\sigma_\Sigma^2=\mathcal O(K)$.

In graph-feedback bandits (GFBs), $\mathcal A=[d]$ consists of $d$ arms and $f_t(a)=\theta_{t,a}$, with $\theta_t\in[0,1]^d$; choosing $a$ reveals its reward and those of its neighbors in a known undirected graph $G$~\citep{caron2012side}. $\alpha(G)$ denotes the independence number, the largest number of pairwise nonadjacent vertices, and $\chi_{\mathrm{cl}}(G)$ denotes the clique-partition number, the minimum number of cliques needed to partition the vertices. These quantities satisfy $\alpha(G)\le\chi_{\mathrm{cl}}(G)$ and describe the graph dependence of the regret bounds~\citep{lykouris2020feedback,lu2021nsgraph}. Thus, the same decision problem encompasses finite and continuous action spaces, as well as feedback that reveals one reward or several outcomes in a round. In all bandit settings, we consider bounded or sub-Gaussian observations, including logistic rewards.

The lion's share of the literature on bandits assumes \emph{stationarity}---i.e., fixed $f_t$, $\theta_t$, $\mathcal P_t$---but this rarely holds in practice due to evolving conditions~\citep{Agrawal:inventory:2019,Cai:adv:2017,Chen:traffic:2020,Lu:adv:2019}.
Non-stationary (NS) settings are often categorized into two types---\emph{gradual drifts} and \emph{abrupt changes}. In the drifting model, $f_t$ and $\mathcal P_t$ evolve slowly under a variation budget constraint~\citep{wei2021master}. In contrast, piecewise stationary (PS) models assume abrupt shifts at unknown change-points, with a total of $N_T$ changes:
\[
 1=:\nu_0<\nu_1<\dots<\nu_{N_T}<\nu_{N_T+1}:=T+1,
\]
with $f_t=f_{t'}$ and $\mathcal P_t=\mathcal P_{t'}$ for $t,t'\in\{\nu_k,\dots,\nu_{k+1}-1\}$. The feedback law is likewise fixed within each segment. Conditional on the segment laws and boundaries, the context and potential feedback are drawn independently across rounds. Item marginals do not depend on which action reveals them; outcomes of different items in the same round may be dependent.

% We assume that, in each stationary segment, $\Pi$ contains a policy that maximizes the mean reward over $\mathcal A$ at every context. Thus the comparator in~\eqref{eq:dynamic_regret} is attainable by one policy throughout that segment. This holds, in particular, for the universal class of mappings from $\mathcal C$ to $\mathcal A$.

NS bandit algorithms are typically either \emph{adaptive}, adjusting continuously, or \emph{restarting}, choosing to unlearn and kickstart the learning process at certain times. They may also be \emph{prior-based} (assuming knowledge of the non-stationarity) or \emph{prior-free}. Prior-based adaptive methods (discounting/sliding window) weigh recent observations more heavily and have been developed for LBs~\citep{cheung2019swlbs,russac2019weighted}, GLBs~\citep{faury2021regboundsnsglbs,russac2020algnsglbs}, SCBs~\citep{russac2021nsweightscbs,wang2023weight}, KBs~\citep{deng2022wgpucb,zhou2021rgpucb-swgpucb}, CSBs~\citep{chen2021nscmab}, and GFBs~\citep{lu2021nsgraph}. Prior-based restarting approaches use budgeted restarts and have been studied for LBs and GLBs~\citep{zhao2020rnsglbs}, KBs~\citep{zhou2021rgpucb-swgpucb}, HBs~\citep{kang2024zooming}, and GFBs~\citep{lu2021nsgraph}.  Detection-based restarting methods are prior-free for LBs and KBs~\citep{hong2023opkb}, CBs~\citep{chen2019adailtcbp}, HBs~\citep{nguyen2025nslipschitz}, CSBs~\citep{zhou2020combinatorial,chen2021nscmab}, and GFBs~\citep{lu2021nsgraph}. 
%CFBs also admit the applicable NS-LB methods through their incidence-vector representation. 
The most closely related work is~\cite{huang2025cdbppsmabs}, which addresses PS multi-armed bandits and introduces techniques that we build upon in establishing our theory.

Among prior-free methods, the black-box paradigm is particularly appealing: it equips a stationary bandit algorithm with non-stationarity handling capabilities without redesigning its learning rule. MASTER~\citep{wei2021master} is a prominent method of this class, detecting non-stationarity by checking for violations of stationary regret bounds. However, its prescribed thresholds prevent detection at practical horizons~\citep{gerogiannis2024blackboxfeas}, while a recently identified error in its regret analysis invalidates the published proof of its theoretical guarantees~\citep{gerogiannis2026darling}. To this end, we consider MASTER empirically but not theoretically. More broadly, methods tailored to individual settings do not by themselves explain how to combine a preferred stationary learner with a practical detector across different action and feedback models. We address this problem with a theoretically grounded, practical black-box framework and comprehensive synthetic and real-world evaluation.

% Our black-box augments a stationary algorithm with any detector satisfying a specific property, which enables flexibility in detector choice. It detects actual changes in the monitored reward means instead of violations of stationary regret bounds. Whether an action reveals one reward, several item outcomes, or neighboring rewards determines which observations are monitored, but not the principle of restarting the learner when a change is detected. Under detectable, sufficiently separated changes, the stationary input's regret transfers to the PS setting whenever the additional exploration cost is of no larger order.

\textbf{Contributions.} We present Detection Augmented Learning (DAL), a \emph{practical} prior-free, black-box detection-based paradigm for general NS bandits. The design is motivated by the need to adapt without prior knowledge of non-stationarity and by a black-box reduction that simplifies non-stationary algorithm design to specifying when to restart a stationary learner. Our method is simple---combining a change detector, a forced exploration schedule, and a stationary bandit algorithm---modular, and easy to implement. Its design applies to PBs, KBs, CBs, HBs, CSBs and CFBs, and GFBs. Theoretically, a single regret theorem transfers the stationary input's guarantee to the PS setting under detectable, sufficiently separated changes. The resulting bounds match the best existing bounds in PS-LB and PS-CB rates and improve all other settings, under the assumptions specified in Table~\ref{table:reg_bounds}. For drifting regimes, we identify conditions for good performance and validate them empirically. Empirically, extensive synthetic and real-world evaluations show consistent gains over both prior-free and prior-based baselines across nine bandit settings. Finally, we introduce a bandit-driven forced-exploration rule for DAL that learns which actions to probe, and show empirically that it can improve performance compared to standard round-robin probing.

\section{The DAL Framework}
\label{sec:dal_framework}
\begin{algorithm}[t]
\small
\caption{\textbf{D}etection \textbf{A}ugmented \textbf{L}earning (\textbf{DAL})}
\label{alg:DAL}
\textbf{Input}: stationary algorithm $\mathcal B$, detector $\mathcal D$, covering set $\mathcal A_{\mathrm e}$, context set $\mathcal C$, frequencies $(\alpha_k)_{k=1}^T$, horizon $T$\\
\textbf{Initialize}: empty monitored histories; $\tau\leftarrow0$, $k\leftarrow1$
\begin{algorithmic}[1]
\FOR{$t=1,\ldots,T$}
 \STATE Observe context $C_t$ and set $i\leftarrow ((t-\tau-1)\bmod\lceil N_{\mathrm e}/\alpha_k\rceil)+1$
 \STATE \textbf{if} $i\le N_{\mathrm e}$ \textbf{then} play $a^{(i)}$ and add the feedback to the corresponding monitored histories
  \STATE $\quad$ \textbf{if} {$\mathcal D$ declares a change on any updated history} \textbf{then}
   \STATE $\quad\quad$ Reset $\mathcal B$ and clear all monitored histories and set
   $\tau\leftarrow t$, $k\leftarrow k+1$
  \STATE \textbf{else} Run $\mathcal B$ for one local step, using its own feedback
\ENDFOR
\end{algorithmic}
\end{algorithm}

The DAL framework is a black-box characterized by a modular structure of three components: a non-stationarity detector, a forced exploration scheme, and a bandit algorithm. We provide high-level ideas of the structure of our approach and formally present our framework in Algorithm~\ref{alg:DAL}.

\textbf{Non-Stationarity Detector.} To identify changes in the environment, DAL uses a general-purpose detector $\mathcal D$ for monitoring shifts in the distribution of judiciously chosen observation sequences obtained through forced exploration. Rather than testing the stationary learner's regret bound, it tests for changes in the means of these observations. We adopt a detector aligned with~\cite{besson2022efficient,huang2025cdbppsmabs}, grounded in the well-established theory of quickest change detection~\citep{vvv_qcd_overview}. With scalar feedback, DAL detects changes in the mean reward associated with a context--action pair, keeping histories $\mathcal H_{(c,a)}$ for $(c,a)\in\mathcal C\times\mathcal A_{\mathrm e}$. With semi-bandit or graph feedback, it instead keeps a history $\mathcal H_j$ for each item or arm, and updates every history revealed by a forced action. Thus, one action may contribute to several detectors. The detector can be replaced without changing the learner, provided that it satisfies the property in Section~\ref{sec:on_detection}.

\textbf{Forced Exploration.} In stationary settings, effective algorithms quickly concentrate on optimal actions for each context, rarely exploring suboptimal actions. In NS environments, however, this behavior may lead to missed changes on these rarely sampled actions, and thus, \emph{forced exploration} is essential for reliable detection. When the action space is large or infinite, exploring all actions becomes infeasible. Therefore, DAL only does extra exploration on a finite \emph{covering set}, $\mathcal A_{\mathrm e}=\{a^{(i)}:i\in[N_{\mathrm e}]\}\subseteq\mathcal A$, in which $a^{(i)}$ denotes the $i$th action in $\mathcal A_{\mathrm e}$. The set is designed such that the mean of at least one monitored history changes whenever a detectable change occurs. Its construction uses both the reward model and the available feedback; for instance, $\lceil d/K\rceil$ feasible actions suffice to observe all $d$ items in top-$K$ selection. In particular, after the $(k-1)$th restart, DAL is forced to play each action in $\mathcal A_{\mathrm e}$ once for $N_{\mathrm e}$ steps, followed by the bandit algorithm for the next $\lceil N_{\mathrm e}/\alpha_k\rceil-N_{\mathrm e}$ steps, repeatedly, until the $k$th restart. Here, $\alpha_k\in(0,1)$ is the exploration frequency, striking a balance between detection delay and regret from extra exploration.

\textbf{Bandit Algorithm.} With a detector $\mathcal D$ and forced exploration, DAL augments a (stationary) bandit algorithm $\mathcal B$: it resets $\mathcal B$ entirely whenever $\mathcal D$ detects a change in any monitored history, and runs $\mathcal B$ with periodic forced exploration otherwise. A key advantage of DAL is its ability to translate strong stationary performance into robust performance under NS conditions. The learner uses its own local clock and observations, so that pausing it for exploration does not alter its stationary interface. Each restart initializes a fresh copy. Any stationary algorithm satisfying the regret requirement of Theorem~\ref{thrm:dal_reg} can be used; retaining an order-optimal rate additionally requires that the stationary input be order-optimal and that exploration not dominate its regret.

\section{Theoretical Insights}
\label{sec:thr_insight}
\subsection{On Effective Detection}
\label{sec:on_detection}

\begin{figure*}[t] % Use figure* to span both columns
\centering
\begin{minipage}{0.48\textwidth}

    \begin{algorithm}[H] % [H] is required to keep it inside the minipage
\caption{\small{\textbf{G}eneralized \textbf{L}ikelihood \textbf{R}atio Test}}
        \label{alg:GLR}
        \begin{algorithmic}[1]
            \small
            \STATE {\bf Input}: History $\mathcal{H}=\{X_1,\dots,X_n\}$, $\delta_\mathrm{F}$, $\delta_\mathrm{D}$, KL divergence $\mathrm{kl}(\cdot,\cdot)$
            \FOR{$k = 1$ to $n-1$}
                \STATE Compute empirical means $\hat{\mu}_{1:k}$, $\hat{\mu}_{k+1:n}$, $\hat{\mu}_{1:n}$
                \STATE $\mathrm{GLR}_k \leftarrow k\,\mathrm{kl}(\hat{\mu}_{1:k}, \hat{\mu}_{1:n})$
                \STATE \quad $\phantom{\mathrm{GLR}_k \leftarrow}\;+ (n-k)\,\mathrm{kl}(\hat{\mu}_{k+1:n}, \hat{\mu}_{1:n})$
                \IF{$\mathrm{GLR}_k \ge \beta_\mathrm{GLR}(n,\delta_\mathrm{F})$}
                    \RETURN \texttt{detection}
                \ENDIF
            \ENDFOR
        \end{algorithmic}
    \end{algorithm}
\end{minipage}
\hfill
\begin{minipage}{0.48\textwidth}
    \begin{algorithm}[H]
        \caption{\small{\textbf{G}eneralized \textbf{S}hiryaev--\textbf{R}oberts Test}}
        \label{alg:GSR}
        \begin{algorithmic}[1]
            \small
            \STATE {\bf Input}: History $\mathcal{H}=\{X_1,\dots,X_n\}$, $\delta_\mathrm{F}$, $\delta_\mathrm{D}$, KL divergence $\mathrm{kl}(\cdot,\cdot)$, $\mathrm{GSR}_{k}\leftarrow0$
            \FOR{$k = 1$ to $n-1$}
                \STATE Compute empirical means $\hat{\mu}_{1:k}$, $\hat{\mu}_{k+1:n}$, $\hat{\mu}_{1:n}$
                \STATE Compute $\mathrm{GLR}_k $ according to Alg. \ref{alg:GLR}
                \STATE $\mathrm{GSR}_k \leftarrow \mathrm{GSR}_k + \exp(\mathrm{GLR}_k)$
                \IF{$\mathrm{GSR}_{k} \ge \beta_\mathrm{GSR}(n,\delta_\mathrm{F})$}
                    \RETURN \texttt{detection}
                \ENDIF
            \ENDFOR
        \end{algorithmic}
    \end{algorithm}
\end{minipage}
\end{figure*}
When selecting a non-stationarity detector, accuracy and efficiency are essential for ensuring optimal regret growth. Any detector aiming to identify distribution shifts inherently requires a certain number of samples, both before and after the change. This sample complexity should scale appropriately to avoid negatively impacting the total regret. %To this end, the GLR and GSR tests achieve a pre- and post-change sample complexity of order $\log T$ for a fixed noise proxy and mean shift~\citep{huang2025sequentialchangedetectionlearning}.

The stopping time $\tau$ of a change detector $\mathcal D$ denotes the time-step at which a change is identified, measured in the detector's own observations. Let $\mathbb P_\nu$ denote the probability law with change-point at $\nu$, and $\mathbb P_\infty$ the law with no change-point. The \emph{latency} $\ell_{\mathcal D}$ is the length of time post-change within which a change is declared with probability at least $1-\delta_{\mathrm D}$, i.e.,
\[
 \ell_{\mathcal D}:=\inf\left\{t\in[T]:\mathbb P_\nu(\tau\ge\nu+t)\le\delta_{\mathrm D},\quad
 \forall\nu\in[m_{\mathcal D}+1,T-t]\right\},
\]
where $m_{\mathcal D}$ is the required pre-change window. A good detector seeks to minimize $\ell_{\mathcal D}$ while ensuring low false-alarm probability over horizon $T$, namely $\mathbb P_\infty(\tau\le T)\le\delta_{\mathrm F}$. We require the following property.

\begin{property}\label{proper:good_CD}
The detector satisfies $\mathbb P_\infty(\tau\le T)\le\delta_{\mathrm F}$, and its pre-change window and latency satisfy
$
 \ell_{\mathcal D}+m_{\mathcal D}=\mathcal O\!\left(\log T+\log(1/\delta_{\mathrm F})+\log(1/\delta_{\mathrm D})\right).
$ We refer to any detector that satisfies this property as a \emph{Horizon-Logarithmic Detector} (HLD).
\end{property}
In \cite{huang2025sequentialchangedetectionlearning} it was shown that the \emph{Generalized Likelihood Ratio} (GLR) and \emph{Generalized Shiryaev-Roberts} (GSR) tests are HLDs for independent sub-Gaussian observations with fixed variance~\citep{huang2025sequentialchangedetectionlearning}. The two HLDs are given in Algorithms \ref{alg:GLR} and \ref{alg:GSR} and $\mathrm{kl}( x;y)$ is KL-divergence between two Gaussian or Bernoulli distributions, depending on the rewards. To select which samples should be fed into the HLD, one needs to properly select the covering set $\mathcal A_{\mathrm e}$, so that it contains actions whose feedback captures changes in the reward function. However, changes cannot be arbitrarily small if detection is to have logarithmic sample complexity with uniform constants. Hence, $\mathcal A_{\mathrm e}$ should be designed such that whenever a change occurs, at least one monitored history exhibits an \emph{appreciable} mean-shift. For scalar-reward histories, define
\begin{equation}
 \Delta_{\mathrm c}:=\min_{k\in[N_T]}\max_{(c,a)\in\mathcal C\times\mathcal A_{\mathrm e}}
 |f_{\nu_k}(c,a)-f_{\nu_k-1}(c,a)|.
 \label{eq:detectability_margin}
\end{equation}
In CSBs and GFBs, the maximum is instead over the monitored item or arm means, giving $\Delta_{\mathrm c}=\min_{k\in[N_T]}\|\theta_{\nu_k}-\theta_{\nu_k-1}\|_\infty$. Then, $\Delta_{\mathrm c}$ captures how well the monitored observations can discern between successive reward functions. Its value crucially affects the performance of the HLD, as their pre- and post-change sample complexity grows with the noise-to-squared-shift ratio~\citep{huang2025cdbppsmabs,huang2025sequentialchangedetectionlearning}. The more discernible the changes are, the easier the detection becomes. Since forced exploration incurs regret, $\mathcal A_{\mathrm e}$ should also be small. We therefore use the structure of the reward function and the available feedback to construct informative covering sets. The following propositions provide such choices; their proofs are given in the Appendix.

\begin{proposition}\label{prop:nspb_ae}
In NS-PBs, $\mathcal A_{\mathrm e}$ can be any maximal linearly independent subset of $\mathcal A$. The same holds in NS-CFBs on the incidence vectors, with $N_{\mathrm e}=r\le d$, the dimension of their span.
\end{proposition}

\begin{proposition}\label{prop:nskb_ae}
In NS-KBs, assume $\mathcal A\subseteq[0,R]^d$, that the permitted RKHS functions are $BL_u$-Lipschitz in the Euclidean norm, and that there is a $\tilde a\in\mathcal A$ such that
$
 \inf_{f\ne f'}|f(\tilde a)-f'(\tilde a)|>L_T,
$
for some $L_T>0$, with the infimum over successive changes. Let $\delta_T=L_T/(2BL_u)$ and let $\mathcal V_T\subseteq\mathcal A$ be the centers of a $\delta_T$-cover. Then $\mathcal A_{\mathrm e}$ can be taken as $\mathcal V_T$. Such a cover can be chosen with $|\mathcal V_T|\le\lceil\sqrt d R/\delta_T\rceil^d$; for $\mathcal A=[0,R]^d$, a regular cell-center grid gives $|\mathcal V_T|\le\lceil\sqrt d R/(2\delta_T)\rceil^d$.
\end{proposition}

\begin{proposition}\label{prop:nsHB_ae}
In NS-HBs, assume that every change satisfies $\|f-f'\|_\infty\ge L_T$ for some $L_T>0$. Let $\delta_T=(L_T/(4L))^{1/\beta}$ and let $\mathcal V_T$ be the centers of a $\delta_T$-cover of $[0,1]^d$ in $\|\cdot\|_\infty$. Then $\mathcal A_{\mathrm e}$ can be taken as $\mathcal V_T$, with $\Delta_{\mathrm c}\ge L_T/2$. The regular grid $\{0,1/M,\ldots,1\}^d$, with $M=\lceil1/(2\delta_T)\rceil$, gives such a cover with $(M+1)^d$ points.
\end{proposition}

\begin{proposition}\label{prop:struct_ae}
In NS-CSBs and NS-GFBs, $\mathcal A_{\mathrm e}$ can be any family of feasible actions whose items or observation neighborhoods, respectively, cover all $d$ items or arms. Every forced cycle then observes each monitored history at least once (items that belong to no feasible action can be discarded). If every reward-function change has sup-norm at least $L_T$, then $\Delta_{\mathrm c}\ge L_T/K$ in CSBs and $\Delta_{\mathrm c}\ge L_T$ in GFBs. A CSB cover always exists with $N_{\mathrm e}\le d$; top-$K$ selection admits $N_{\mathrm e}=\lceil d/K\rceil$, but this smaller size need not hold for a general feasible family. In GFBs, a greedy maximal independent set is a valid dominating set, with $N_{\mathrm e}\le\alpha(G)$.
\end{proposition}

\begin{remark}\label{rem:nscb_ae}
In NS-CBs, if $f_t$ and $\mathcal A$ satisfy the preceding structural assumptions simultaneously for all contexts, we can set $\mathcal A_{\mathrm e}$ similarly. Otherwise, set $\mathcal A_{\mathrm e}=\mathcal A$, as $\mathcal A$ is finite.
\end{remark}

The constructions in Propositions~\ref{prop:nspb_ae}--\ref{prop:struct_ae} and Remark~\ref{rem:nscb_ae} are one-time preprocessing steps performed before DAL is run, and do not introduce per-round overhead. In PBs and CFBs, the cost scales with the number of inspected candidate actions and the corresponding linear-independence checks, rather than with the horizon $T$. For continuous actions, the resolution of a fixed cover determines which changes are detectable. The corresponding propositions can be read as conditions on the environments covered by that choice, setting the resolution using the unknown true minimum change would instead require a prior. Spanning or covering alone does not ensure a uniform statistical margin, so the regret analysis retains the requirement $\Delta_{\mathrm c}>0$.

\textbf{Prior-Free Tuning.} With the HLD considered here, DAL does not require $N_T$, the change-points $\nu_k$, the minimum change size, or the segment lengths as inputs. Its exploration frequency uses the observed restart index $k$, and the HLD (GLR/GSR) thresholds use confidence levels and a known noise bound, not the unknown pre- or post-change means. In continuous settings, $\mathcal A_{\mathrm e}$ is fixed independently of the future changes. The quantities $L_T$, $\Delta_{\mathrm c}$, $m_{\mathcal D}$ and $\ell_{\mathcal D}$ describe the detectable environments and the separation needed for the guarantee; their values are not supplied to DAL. Thus DAL is prior-free with respect to non-stationarity, while its guarantee remains conditional on detectable, sufficiently separated changes.

\subsection{On Order-Optimality in Piecewise Stationary Environments}
\label{sec:order_opt}
In the PS setting, restarting a stationary learner at the true change-points gives a natural benchmark. DAL aims to attain the same regret order without knowing their number or locations. As in the detection-based analysis of~\cite{huang2025cdbppsmabs}, the intervals between change-points must be long enough to acquire the samples needed for detection. For brevity, we first define the relevant quantities and then state the assumption. For the contextual guarantee, assume a common finite support $\mathcal C$, with $\mathcal P_t(c)>0$ for all $c\in\mathcal C$ and $t\in[T]$.

\begin{definition}\label{def:sep_scaling}
For the settings with a fixed context, i.e., PS-PBs, PS-KBs, PS-HBs, PS-CSBs, PS-CFBs and PS-GFBs, let
$
 m_k\coloneqq\left\lceil\frac{N_{\mathrm e}}{\alpha_k}\right\rceil m_{\mathcal D}, \: \ell_k\coloneqq\left\lceil\frac{N_{\mathrm e}}{\alpha_k}\right\rceil\ell_{\mathcal D}.
$
For PS-CBs, let
$
m_k\coloneqq\left\lceil\frac{N_{\mathrm e}}{\alpha_k}\right\rceil
 \left\lceil\frac{m_{\mathcal D}}s+\frac{\log T}{4s^2}
 +\sqrt{\frac{m_{\mathcal D}\log T}{2s^3}+\frac{(\log T)^2}{16s^4}}\right\rceil,
 \ell_k\coloneqq\left\lceil\frac{N_{\mathrm e}}{\alpha_k}\right\rceil
 \left\lceil\frac{\ell_{\mathcal D}}s+\frac{\log T}{4s^2}
 +\sqrt{\frac{\ell_{\mathcal D}\log T}{2s^3}+\frac{(\log T)^2}{16s^4}}\right\rceil,
$
for $k\in[N_T]$, with $s\coloneqq\min_{c\in\mathcal C,t\in[T]:\mathcal P_t(c)>0}\mathcal P_t(c)$.
\end{definition}

\begin{assumption}\label{assum:cp_assum}
For $N_T\ge1$, assume $\nu_1>m_1$ and $\nu_k-\nu_{k-1}\ge\ell_{k-1}+m_k$ for $k\in\{2,\ldots,N_T\}$.
\end{assumption}

In the fixed-context settings, DAL performs round-robin forced exploration every $\lceil N_{\mathrm e}/\alpha_k\rceil$ rounds, so each history receives at least one observation per cycle. In PS-CBs, a context--action pair is observed only in expectation, at least once every $\lceil N_{\mathrm e}/\alpha_k\rceil/s$ rounds. The larger windows in Definition~\ref{def:sep_scaling} ensure enough pre- and post-change samples with high probability. Assumption~\ref{assum:cp_assum} is used for the guarantee, but is not imposed in the experiments.

Based on Algorithm~\ref{alg:DAL}, DAL can incorporate any stationary bandit algorithm. Since different algorithms yield different regret guarantees, its regret depends on that of its input. To characterize DAL's performance under piecewise stationarity, we employ the methodology of~\cite{huang2025cdbppsmabs}, incorporating the regret analysis of the stationary bandit algorithm and that of the change detector. The full proof is given in Appendix~\ref{app:thm_proof}.
\begin{theorem}\label{thrm:dal_reg}
For the PS setting, consider DAL with an HLD satisfying Property~\ref{proper:good_CD}, a stationary bandit algorithm $\mathcal B$ with the same concave, increasing expected-regret upper bound $R_{\mathcal B}(n)$ for every local prefix length $n\le T$ in every segment, and a covering set $\mathcal A_{\mathrm e}$ with $\Delta_{\mathrm c}>0$ whenever $N_T\ge1$. If Assumption~\ref{assum:cp_assum} holds, set
$
 \alpha_k=\frac{\sqrt{k|\mathcal C|N_{\mathrm e}}}{2\sqrt T\log^2T},\:\delta_{\mathrm F}=\delta_{\mathrm D}=T^{-\eta},\eta>1,
$
then, DAL's regret satisfies
$
 R_T=\widetilde{\mathcal O}\!\left(
 (N_T+1)R_{\mathcal B}\!\left(\frac{T}{N_T+1}\right)
 +\bar\Delta\sqrt{|\mathcal C|N_{\mathrm e}(N_T+1)T}\right),
$
where $\bar\Delta$ is the maximum mean-reward gap. For CSBs and GFBs, set $\delta_{\mathrm F}=T^{-\eta}/d$ instead.
\end{theorem}

% The first term collects the stationary learning cost; the second accounts for forced exploration, detection delay and failures. Monitoring $d$ items or arms changes the false-alarm union bound, but not the number $N_{\mathrm e}$ of forced actions. The constants may depend on the monitored noise-to-squared-shift ratio and on $s$; the explicit terms are retained in the proof. When $N_T=0$, no separation or detection-margin condition is needed.

Using Theorem~\ref{thrm:dal_reg}, Propositions~\ref{prop:nspb_ae}--\ref{prop:struct_ae} and Remark~\ref{rem:nscb_ae}, we present the regret of DAL.
\begin{corollary}\label{corr:dal_optimal}
Assume the conditions of Theorem~\ref{thrm:dal_reg}. In PBs, KBs, CBs, CSBs, CFBs and GFBs, suppose $R_{\mathcal B}(n)=\widetilde{\mathcal O}(C_{\mathcal B}\sqrt n)$ and select the corresponding covering set with $\bar\Delta\sqrt{|\mathcal C|N_{\mathrm e}}=\mathcal O(C_{\mathcal B})$. Then DAL attains
$
 R_T=\widetilde{\mathcal O}\!\left(C_{\mathcal B}\sqrt{(N_T+1)T}\right).
$
In HBs, a fixed cover and an input with $R_{\mathcal B}(n)=\widetilde{\mathcal O}(n^{(d_z+1)/(d_z+2)})$ give
$
 R_T=\widetilde{\mathcal O}\!\left((N_T+1)^{1/(d_z+2)}T^{(d_z+1)/(d_z+2)}\right).
$
\end{corollary}

That is, if the stationary algorithm has order-optimal regret and its learning term dominates exploration, DAL retains its rate under the stated detection conditions. Corollary~\ref{corr:dal_optimal} yields regret bounds across all considered settings while allowing flexible choices of the stationary algorithm. Choosing the stationary algorithms in Section~\ref{section:practical_perf}, which achieve optimal stationary regret, leads to the bounds summarized in Table~\ref{table:reg_bounds}. 

\begin{table*}[t]
\centering\small
\caption{Regret comparisons for PS bandits, under Assumption~\ref{assum:cp_assum}. ``$\bullet$'' denotes prior-based tuning. DAL attains the best regret or improves the best existing bounds (\dalimprove{blue} = improvement).}
\label{table:reg_bounds}
\renewcommand{\arraystretch}{1.18}
\setlength{\tabcolsep}{4pt}
\begin{tabular}{l|c|c|c}
\hline
\textbf{Setting}&\shortstack{\textbf{Input Regret}}&\shortstack{\textbf{Best Existing Regret}}&\shortstack{\textbf{DAL Regret}}\\
\hline
PS-LBs & $\tilde{\mathcal{O}}(d\sqrt T)$ \cite{abbasi-yadkori2011linucb}& $\tilde{\mathcal{O}}(d\sqrt{N_TT})$ \cite{hong2023opkb} & $\tilde{\mathcal{O}}(d\sqrt{N_TT})$\\
\hline
PS-GLBs & $\tilde{\mathcal{O}}(d\sqrt T)$ \citep{filipi2010genlb}
& $\tilde{\mathcal{O}}(d^{2/3}N_T^{1/3}T^{2/3})^\bullet$ \cite{wang2023weight} & $\dalimprove{\tilde{\mathcal{O}}(d\sqrt{N_TT})}$\\
\hline
PS-SCBs & $\tilde{\mathcal{O}}(d\sqrt T)$ \citep{lee2025ofuglb}
 & $\tilde{\mathcal{O}}(d^{2/3}N_T^{1/3}T^{2/3})^\bullet$ \cite{wang2023weight}& $\dalimprove{\tilde{\mathcal{O}}(d\sqrt{N_TT})}$\\
\hline
PS-KBs & $\tilde{\mathcal{O}}(\sqrt{\gamma_TT})$ \citep{salgia2024reds}
& $\tilde{\mathcal{O}}(\sqrt{d\gamma_TN_TT})$ \cite{hong2023opkb}& $\dalimprove{\tilde{\mathcal{O}}(\sqrt{\gamma_TN_TT})}$\\
\hline
PS-CBs & $\tilde{\mathcal{O}}(\sqrt{|\mathcal A|T\log|\Pi|})$ \citep{foster2020squarecb}
& $\tilde{\mathcal{O}}(\sqrt{|\mathcal A|N_TT\log|\Pi|})$ \cite{chen2019adailtcbp} & $\tilde{\mathcal{O}}(\sqrt{|\mathcal A|N_TT\log|\Pi|})$\\
\hline
PS-HBs (General $d,\beta$) & $\tilde{\mathcal{O}}(T^{\frac{d+\beta}{d+2\beta}})$ \citep{bubeck2011xarmed}
& $\tilde{\mathcal{O}}(N_T^{\frac{\beta}{d+3\beta}}T^{\frac{d+2\beta}{d+3\beta}})^\bullet$ \cite{kang2024zooming} & $\dalimprove{\tilde{\mathcal{O}}(N_T^{\frac{\beta}{d+2\beta}}T^{\frac{d+\beta}{d+2\beta}})}$\\
PS-HBs (common $d_z$) & $\tilde{\mathcal{O}}(T^{\frac{d_z+1}{d_z+2}})$ \citep{bubeck2011xarmed}
& $\tilde{\mathcal{O}}(N_T^{\frac1{d_z+3}}T^{\frac{d_z+2}{d_z+3}})^\bullet$ \cite{kang2024zooming} & $\dalimprove{\tilde{\mathcal{O}}(N_T^{\frac1{d_z+2}}T^{\frac{d_z+1}{d_z+2}})}$\\
PS-HBs ($d=1$ or $\beta=1$) & $\tilde{\mathcal{O}}(T^{\frac{d+\beta}{d+2\beta}})$ \citep{bubeck2011xarmed}
& $\tilde{\mathcal{O}}(N_T^{\frac{\beta}{d+2\beta}}T^{\frac{d+\beta}{d+2\beta}})$ \cite{nguyen2025nslipschitz}& $\tilde{\mathcal{O}}(N_T^{\frac{\beta}{d+2\beta}}T^{\frac{d+\beta}{d+2\beta}})$\\
\hline
PS-CSBs (General) & $\tilde{\mathcal{O}}(K\sqrt{dT})$ \citep{chen2013cmab}
& $\tilde{\mathcal{O}}(K\sqrt{dN_TT})$ \cite{chen2021nscmab}& $\tilde{\mathcal{O}}(K\sqrt{dN_TT})$\\
PS-CSBs (Top-$K$) & $\tilde{\mathcal{O}}(\sqrt{KdT})$ \cite{kveton2015tight}& $\tilde{\mathcal{O}}(K\sqrt{dN_TT})$ \cite{chen2021nscmab}& $\dalimprove{\tilde{\mathcal{O}}(\sqrt{KdN_TT})}$\\
\hline
PS-CFBs & $\tilde{\mathcal{O}}(\sqrt{KdrT})$ \citep{abbasi-yadkori2011linucb}
& $\tilde{\mathcal{O}}(\sqrt{KdrN_TT\log|\mathcal A|})$ \cite{hong2023opkb}& $\dalimprove{\tilde{\mathcal{O}}(\sqrt{KdrN_TT})}$\\
\hline
PS-GFBs & $\tilde{\mathcal{O}}(\sqrt{\alpha(G)T})$ \citep{lykouris2020feedback}& $\tilde{\mathcal{O}}(\sqrt{\chi_{\mathrm{cl}}(G)N_TT})$ \cite{lu2021nsgraph}& $\dalimprove{\tilde{\mathcal{O}}(\sqrt{\alpha(G)N_TT})}$\\
\hline
\end{tabular}
\end{table*}

\subsection{On Drifting Environments}
\label{sec:drift}
\begin{figure*}[t]
 \centering
 \includegraphics[width=0.8\linewidth]{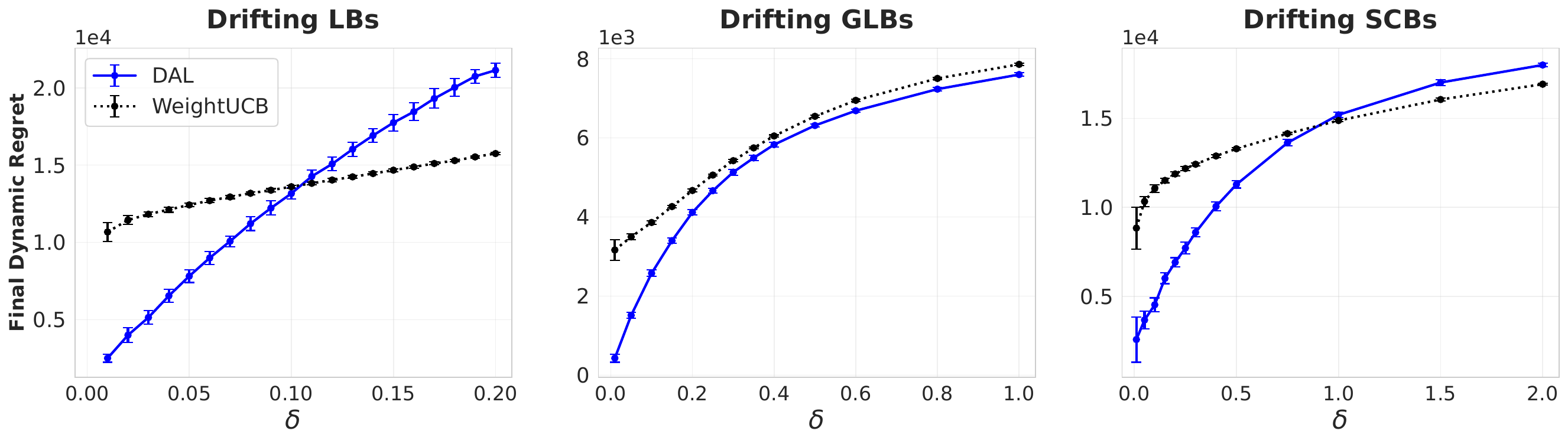}
 \caption{Final dynamic regret vs.\ radius $\delta$ of the drift, mean $\pm$ one standard deviation over $30$ runs: DAL vs.\ WeightUCB; the range of $\delta$ differs between panels. DAL performs best at small drift radii; WeightUCB overtakes DAL beyond $\delta\approx0.1$ in LBs and $\delta\approx1$ in SCBs.}
 \label{fig:final_reg_vs_delta_drift}
\end{figure*}

While our results in the previous section focused on PS environments, naturally we ask whether DAL performs well in drifting environments. As a first step to study this, we perform an experiment with LBs, GLBs and SCBs, the three parametric settings covered by WeightUCB. Specifically, the parameter $\theta_t$ at each time-step evolves according to a projected random walk,
$
 \theta_{t+1}:=\operatorname{Proj}_{\{\theta:\|\theta\|\le S\}}
 (\theta_t+\zeta_{t+1}),
$
where $\zeta_{t+1}\in\mathbb R^d$ is chosen uniformly over a $\delta$-ball. We compare the cumulative dynamic regret up to time $T$ of DAL with GLR and the corresponding stationary learner, and WeightUCB, over a range of $\delta$'s in Figure~\ref{fig:final_reg_vs_delta_drift}. WeightUCB receives the true variation of each run. The LB experiment uses $d=5$ and $\varepsilon_t\sim\mathcal N(0,0.1)$; the remaining parameters, and those of GLBs and SCBs, are as in Section~\ref{section:practical_perf}. We use $30$ runs per point.

We now shed light on our hypothesis behind the observations from Figure~\ref{fig:final_reg_vs_delta_drift}. Write $h_{t+1}:=\theta_{t+1}-\theta_t$ for the actual increment after projection. Since projection is nonexpansive and $\theta_t$ lies in the parameter ball, $\|h_{t+1}\|\le\delta$. In LBs, for playing an action $a\in\mathcal A$ at time $t+1$, we get the random reward
$
 X_{t+1}=\langle\theta_t,a\rangle+\langle h_{t+1},a\rangle+\varepsilon_{t+1}.
$
If $\theta_t$ does not change, then the reward from playing action $a$ would have been $X'_{t+1}=\langle\theta_t,a\rangle+\varepsilon'_{t+1}$, where $\varepsilon'_{t+1}$ is another realization of the noise. When $\delta$ is small relative to the observation noise, the small mean-shift $\langle h_{t+1},a\rangle$ may be difficult to distinguish from noise over a short window. The same reasoning applies to GLBs and SCBs: if the link has Lipschitz constant $L_\mu$ on the bounded parameter domain, the per-round mean shift is at most $L_\mu\|a\|\delta$.

For practical purposes, a slowly drifting environment may therefore resemble a stationary one over the learner's recent observations. As a result, one expects an algorithm tailored to handle piecewise stationarity to perform reasonably well for slowly drifting environments, restarting when the accumulated mean shift becomes appreciable. Conversely, if $\delta$ is large, the accumulated variation can invalidate the stationary learner's estimates before a reliable restart is possible. This is a statistical explanation for the observed behavior, not an application of the PS theorem to drift: the effective variation need not be zero-mean, and the detector's piecewise-stationary latency guarantee does not by itself cover an arbitrary random walk.

The experiments support this explanation across the three reward models. DAL performs better than WeightUCB for smaller values of $\delta$, but the conclusion reverses upon increasing $\delta$ in LBs and SCBs. At the smallest tested radii, WeightUCB incurs $7.3$ times DAL's regret in GLBs and $3.4$ times in SCBs. In SCBs, WeightUCB overtakes DAL beyond $\delta\approx1$, with parameter radius $S=3$. In GLBs, it approaches DAL but remains $3$--$4\%$ above it up to $\delta=1$. Thus, small per-round variation favors DAL in these experiments, whereas a weighted method that tracks drift can overtake it as the variation grows.

\section{Practical Performance}
\label{section:practical_perf}

\subsection{Experimental Baselines}
We evaluate the relevant prior-free and prior-based methods referenced in Section~\ref{sec:intro}, highlighting the strongest state-of-the-art algorithms applicable to PS and drifting settings across both synthetic and real-world benchmarks. These include ADA-OPKB~\citep{hong2023opkb} for LBs and KBs, ADA-ILTCB+~\citep{chen2019adailtcbp} for CBs, ZoomingTS-R~\citep{kang2024zooming} and MDBE~\citep{nguyen2025nslipschitz} for HBs, GLR-CUCB~\citep{zhou2020combinatorial}, CUCB-SW and ADA-LCMAB~\citep{chen2021nscmab} for CSBs, and UCB-NEW, SEASIDE and ASG~\citep{lu2021nsgraph} for GFBs. In CFBs, the incidence-vector representation permits applying RestartUCB~\citep{zhao2020rnsglbs}, SW-LinUCB~\citep{cheung2019swlbs}, D-LinUCB~\citep{russac2019weighted}, and MASTER~\citep{wei2021master}. We also compare against the prior-based discounted approaches WeightUCB~\citep{wang2023weight} in parametric models and WGP-UCB~\citep{deng2022wgpucb} in KBs, together with the applicable sliding-window and budget-restart methods.

Unless otherwise stated, all methods are tuned according to their original works. To maintain readability, we group algorithms by paradigm (discounted, sliding-window, budget-restart) while keeping distinct methods separate when they differ meaningfully. In real-world experiments, we focus on the strongest methods from the corresponding synthetic comparisons.

\subsection{Practical Tuning of DAL}
\textbf{Detector and Bandit Selection.} Across all settings, DAL uses the \emph{Generalized Likelihood Ratio (GLR)} and the \emph{Generalized Shiryaev--Roberts (GSR)} tests~\citep{huang2025sequentialchangedetectionlearning} as the detector $\mathcal D$. We use the practical thresholds
$
 \beta_{\mathrm{GLR}}(n,\delta_{\mathrm F})=\log(n^{3/2}/\delta_{\mathrm F}),\:\beta_{\mathrm{GSR}}(n,\delta_{\mathrm F})=n^{5/2}/\delta_{\mathrm F},
$
with $\delta_{\mathrm F}=T^{-1/2}$, following~\cite{huang2025cdbppsmabs,besson2022efficient}. Concretely, Gaussian GLR and GSR pair with LinUCB~\citep{abbasi-yadkori2011linucb} in LBs, GLM-UCB~\citep{filipi2010genlb} in GLBs, REDS~\citep{salgia2024reds} in KBs, HOO~\citep{bubeck2011xarmed} in HBs, CUCB~\citep{chen2013cmab} in CSBs, LinUCB on incidence vectors in CFBs, and UCB-N~\citep{caron2012side} in GFBs. Bernoulli GLR and GSR pair with OFUGLB~\citep{lee2025ofuglb} in SCBs and SquareCB~\citep{foster2020squarecb} in CBs. Unless stated otherwise, we use the exploration schedule in Theorem~\ref{thrm:dal_reg}, $\alpha_k=\sqrt{k|\mathcal C|N_{\mathrm e}}/({2\sqrt T\log^2T}).$

\textbf{Construction of $\mathcal A_{\mathrm e}$.}
For PBs and CFBs, we follow Proposition~\ref{prop:nspb_ae}: we greedily select linearly independent action vectors until they span all feasible action vectors, using incidence vectors in CFBs. In CFBs, merely covering all items is insufficient because item changes may cancel in the observed sums; spanning the incidence vectors prevents a reward-function change from leaving every monitored mean unchanged. For continuous KBs and HBs, we use cover centers as in Propositions~\ref{prop:nskb_ae} and~\ref{prop:nsHB_ae}; in HBs, a regular grid ensures that every action lies near a monitored point. The finite KB action sets in our experiments are monitored in full. In CBs, we take $\mathcal A_{\mathrm e}=\mathcal A$, as noted in Remark~\ref{rem:nscb_ae}. For CSBs and GFBs, we follow Proposition~\ref{prop:struct_ae}. In CSBs, we repeatedly add a feasible action containing an uncovered item until every active item is covered; this suffices because item outcomes are observed separately. In GFBs, we construct a greedy maximal independent set by repeatedly selecting an uncovered vertex and marking it and its neighbors as covered until every vertex is covered. The resulting set is dominating, so every arm is observed during each completed forced cycle. These are one-time preprocessing steps performed independently of the unknown changes. 

\textbf{Practical Variant.} In the finite-action experiments, we also monitor the observations obtained by $\mathcal B$, using reward histories for actions or histories for individual items and neighboring arms. This often detects changes before a forced cycle is completed. Theorem~\ref{thrm:dal_reg} analyzes the forced-only histories of Algorithm~\ref{alg:DAL}; the all-observation implementation is the practical variant which improves performance. This does not affect the theoretical properties of the algorithm, and we employ this variation for our experiments.

\subsection{Synthetic Experiments}
\label{sec:synth_exps}
We evaluate DAL in nine bandit settings under piecewise stationarity and drifting non-stationarity. The two HB instances and the two CSB action families give eleven benchmarks in total. These experiments vary the reward model, action space, available feedback, and frequency of change, while retaining the same detector-and-restart design.

\textbf{Common Parameters.} We use $T=50{,}000$ and average results over $15$ independent runs. In PBs, KBs and CBs, the action set contains $100$ actions of dimension $10$, generated as described below, and the contexts in CBs also have dimension $10$. The HB, combinatorial and graph benchmarks have structured action sets (grids, combinatorial families and a graph), which we describe together with their benchmarks below.

\textbf{PBs.} Actions are sampled from $\mathcal N(0,I)$ and rescaled to an $L$-ball. We use $S=L=1$ for LBs and GLBs and $(L,S)=(1,3)$ for SCBs. At each initialization or change, entries of $\theta_t$ are drawn i.i.d.\ $\mathrm{Unif}[-1,1]$ and then rescaled to satisfy $\|\theta_t\|\le S$. The GLB and SCB link is $\mu(x)=(1+e^{-x})^{-1}$. Noise in LBs and GLBs is $\varepsilon_t\sim\mathcal N(0,0.01)$ per round, and the SCB rewards are sampled as $\mathrm{Bernoulli}(\mu(\langle\theta_t,A_t\rangle))$.

\textbf{KBs.} We use the squared-exponential kernel with length scale $0.2$ and noise $\mathcal N(0,0.01)$. Following~\cite{deng2022wgpucb}, reward functions are generated as finite kernel expansions,
$f(a)=\sum_{i=1}^{200}\alpha_i k(a,x_i)$, with $\alpha_i\sim\mathrm{Unif}[-1,1]$. The covering set is the full finite action set.

\textbf{CBs.} The contexts $C_t\in\mathbb R^{d_c}$ with $d_c=10$ are randomly drawn each round from a fixed pool of $1000$ normalized vectors. At every initialization or change, at least one of $f_t$ or $\mathcal P_t$ changes. For $a\in\mathcal A$, define the $[0,1]$-clipped mean reward
$
 f_t(C_t,a)=\Big[b_a+z^{(\sigma)}\sigma(u_a^\top C_t)
 +z^{(s)}\sin(v_a^\top C_t)+z^{(x)}C_{t,2}C_{t,3}\Big]_{[0,1]},
$
where $u_a,v_a\sim\mathcal N(0,I)$, $b_a\sim\mathrm{Unif}[0.3,0.7]$, and $z^{(\sigma)},z^{(s)},z^{(x)}$ are drawn uniformly from $[0.25,0.45]$, $[0.15,0.35]$ and $[0.10,0.25]$, respectively. Rewards are sampled as $\mathrm{Bernoulli}(f_t(C_t,A_t))$. Since there is no arm-related structure, we set $\mathcal A_{\mathrm e}=\mathcal A$ (Remark~\ref{rem:nscb_ae}).

\textbf{HBs.} Two benchmarks use $(d,\beta)\in\{(2,1),(1,1/2)\}$ and
$f_t(a)=0.9-0.65\|a-\theta_t\|_\infty^\beta$.
Under abrupt changes, $\theta_t$ is redrawn uniformly from $[0.1,0.9]^d$. Under drift, it moves linearly from $(0.22,0.24)$ to $(0.78,0.76)$, using the first coordinate for $d=1$, with a common offset drawn from $\mathrm{Unif}[-0.015,0.015]^d$. HOO operates on $121$ grid points in one dimension or a $21\times21$ grid in two dimensions. The detection cover has nine regularly spaced probes. Observation noise is $\mathcal N(0,0.1^2)$.

\textbf{CSBs.} In the combinatorial and graph benchmarks, the item or arm means are permutations of $d$ equally spaced values in $[0.08,0.84]$. Every abrupt change draws a new permutation; drift interpolates from the initial permutation to its cyclic rotation by $\lfloor d/3\rfloor$. Top-$K$ selection uses $d=12,K=3$ and four covering actions. Shortest paths use a layered graph with three nodes per layer, $15$ edges and nine paths of length three. Each path includes a distinct middle edge, so all nine paths are necessary to cover the edges. Action optimization is exact by enumeration. Since all paths have equal length, maximizing reward is equivalent to minimizing edge costs $1-X_{t,j}$. Semi-bandit item noises have variance $0.1^2$.

\textbf{CFBs.} The full-bandit benchmark has $d=8,K=3$ and all $56$ feasible triples, with item means generated as above. Its eight covering actions are $\{012,013,014,015,016,017,023,123\}$ on items $\{0,\ldots,7\}$; their incidence vectors span $\mathbb R^8$. Independent item noises have variance $0.1^2/3$, so the observed sum has variance $0.1^2$.

\textbf{GFBs.} The feedback graph is a self-observing $12$-cycle, with means and drift as above. Each action reveals its own observation and its two neighbors. The cover $\{0,3,6,9\}$ is a maximal independent set and is minimal as a dominating set: one action observes three vertices. Here $\alpha(G)=6$, and a partition into six edges gives $\chi_{\mathrm{cl}}(G)=6$. Noise has variance $0.1^2$.

\textbf{Piecewise Stationarity.} We adopt the geometric change-point model~\citep{gerogiannis2024blackboxfeas}, and sample the intervals between the change-points according to a geometric distribution with parameter $\rho=T^{-\xi}$, for $\xi\in\{0.4,0.6,0.8\}$. The expected number of changes is approximately $660$, $76$ and $8.7$, respectively. Parameters are redrawn independently at each change, according to the construction of each benchmark. We do not impose any restriction on the lengths of the intervals between change-points in our experiments.

\textbf{Drifting Non-Stationarity.} We drift linearly over $T$ rounds from an initial value to a final value, where the end-points are chosen as in the beginning of the section. Specifically, $\theta_t=(1-t/T)\theta_{\rm init}+(t/T)\theta_{\rm final}$ in parametric, combinatorial and graph benchmarks; in HBs the same rule moves the peak. In KBs, we interpolate between $f_{\rm init}$ and $f_{\rm final}$, and in CBs between the coefficients defining the two reward functions.

\begin{figure}
\centering
\includegraphics[height=0.71\textheight,width=\linewidth,keepaspectratio]{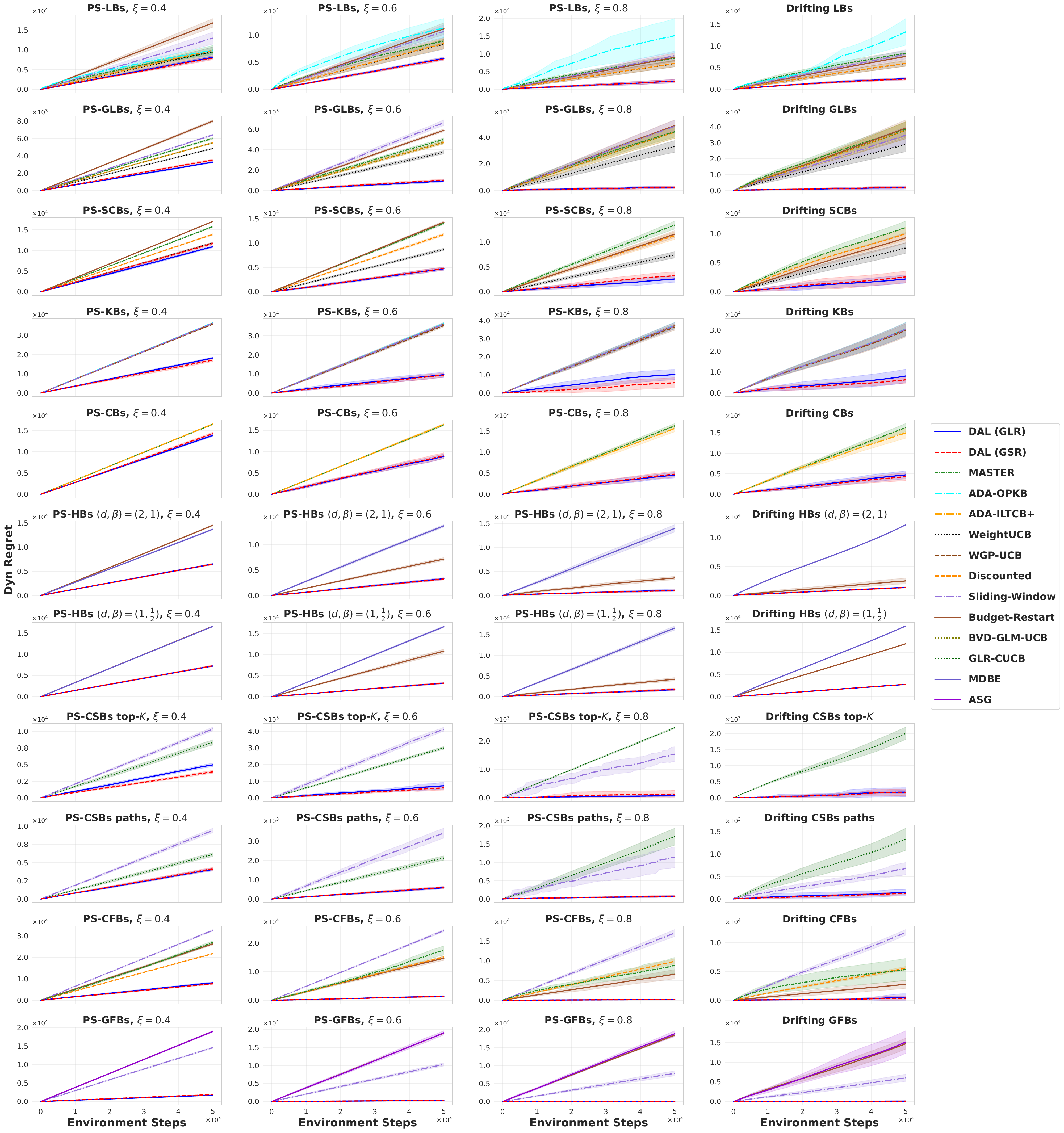}
\caption{Dynamic regret in all synthetic benchmarks (lower is better), averaged over $15$ runs. Rows correspond to the bandit settings and action families; the four columns use geometric change-points with $\xi=0.4,0.6,0.8$ and drift, respectively.}
\label{fig:synthetic}
\end{figure}

\textbf{Experimental Results.} Figure~\ref{fig:synthetic} presents the dynamic regret across all settings and regimes. DAL follows its stationary learner between forced rounds and restarts learning when a change detector flags a mean shift in the monitored observations. Unlike budgeted restarting, it need not discard observations simply because a prescribed block has ended. Both GLR and GSR give strong performance across the reward and feedback models, including under drift. The gains are largest for long stationary segmentsm, while with frequent changes, more segments are missed and the advantage narrows, but persists. Regarding drifting non-stationarity, DAL performs well despite the absence of stationary segments. None of these experiments enforces the separation condition of the theory.

\subsection{Real-World Experiments}
\label{sec:real-world-exp}
To validate DAL's practical effectiveness, we conduct experiments on nine real-world datasets spanning microarchitecture, financial time series, clinical-trial records, click logs, and sensor measurements. The constructions follow exactly \cite{gerogiannis2026dal}. Each comparison is averaged over $15$ independent runs.

\textbf{Micro-Prefetcher Selection.}
We evaluate on an NS bandit benchmark constructed from the microarchitecture prefetcher dataset of \cite{gerogiannis2023bandit}. The environment has $K=11$ prefetcher configurations (actions) with rewards given by normalized IPC in $[0,1]$ over a horizon $T=26224$.
Following \cite{gerogiannis2023bandit}, we also evaluate D-UCB \cite{kocsis2006discounted} in its native form, while all other methods are modeled as NS-SCBs.
Performance is measured by cumulative reward.

\textbf{Stock Market Benchmarks.}
We consider two NS-KB benchmarks following \cite{deng2022wgpucb}: one based on their original Quandl dataset and one constructed from stocks retrieved via Yahoo Finance.\footnote{Data retrieved from Yahoo Finance using the publicly available \texttt{yfinance} package. Used solely for non-commercial, academic research purposes.}
In the Yahoo-based benchmark, we retain assets with sufficient history ($T=2000$) and select the $50$ most volatile as actions.
Rewards are derived from daily closing prices, the empirical price covariance is used as the kernel, and we add i.i.d. $\mathcal{N}(0,0.01)$ noise to increase difficulty.
Evaluation is by dynamic regret.

\textbf{COVID-NMA Clinical Benchmark.}
We construct an NS-SCB benchmark from the open COVID-NMA database \citep{boutron2025coviddata}, modeling treatment effectiveness over time using a union endpoint on reported outcomes.
Treatments are mapped into $K=13$ classes and month-level counts are expanded and concatenated into $T\approx 7.4\times 10^4$.
Performance is evaluated by cumulative reward.

\textbf{Click Log Benchmarks.}
We evaluate on the Yahoo! R6A click log dataset\footnote{Yahoo! Front Page Today Module User Click Log Datasets: \url{https://webscope.sandbox.yahoo.com}.}, yielding an NS-SCB problem with $K=64$ actions and horizon $T=50000$; the metric is dynamic regret.
In addition, we construct a fixed-arm replay benchmark from the Yahoo! R6B click logs as an NS-CB with $51$ actions and $T=50000$, following the logged-bandit evaluation approach of~\cite{li2011unbiasedcontext}; the metric is replay cumulative reward.

\textbf{Live Traffic Benchmark.}
We build an NS-GLB benchmark from the Criteo live traffic dataset~\citep{diemert2017criteodata}, following \cite{russac2019weighted} but introducing changes via a geometric change-point model and using a horizon of $T=50000$. Evaluation is based on dynamic regret.

\textbf{Sensor Correlation Benchmark.}
We use the Bioliq sensor dataset of \cite{komiyama2024globalnsmab} to construct an NS-SCB environment with $K=190$ actions over a week-long horizon.
Rewards are binary indicators derived from temporal correlation statistics following \cite{komiyama2024globalnsmab}.
Performance is measured by cumulative reward.

\textbf{Ad Recommendation Benchmark.}
We evaluate on the Zozo ad recommendation environment from \cite{komiyama2024globalnsmab} but retaining all $K=80$ ads as actions.
Rewards are binary click indicators and evaluation is by cumulative reward.

\begin{figure}
\centering
\includegraphics[width=0.8\linewidth]{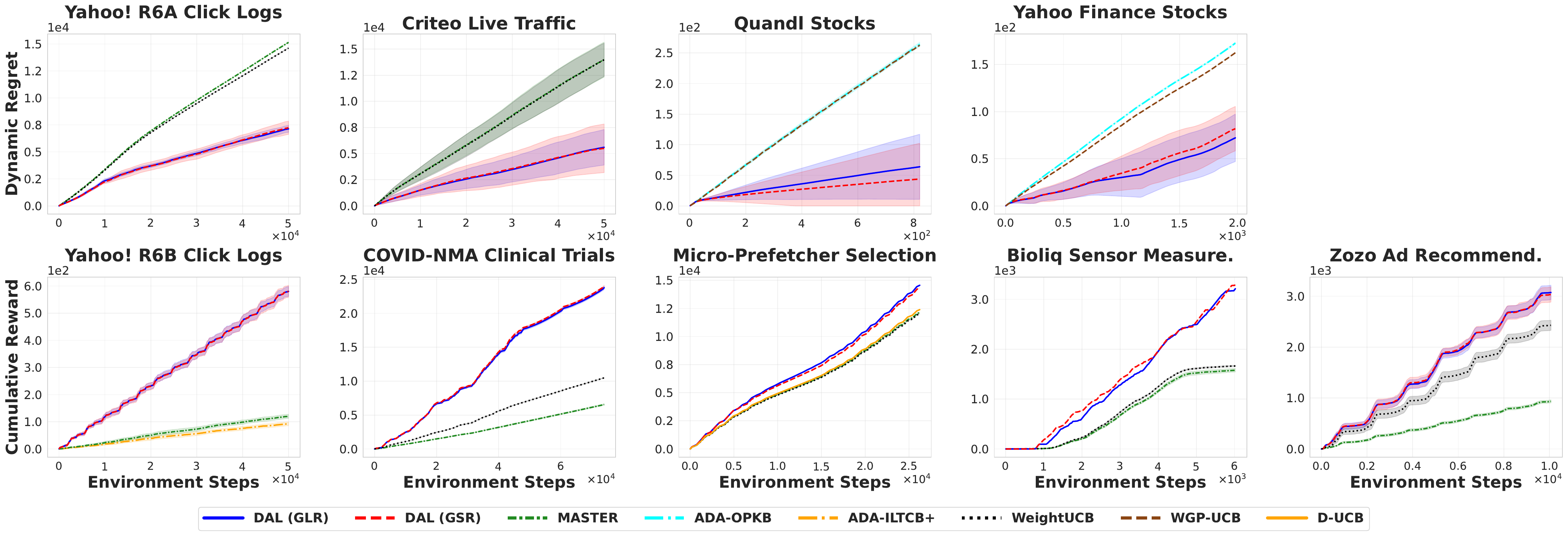}
\caption{Real-world results, averaged over $15$ independent runs. Top: dynamic regret (lower is better). Bottom: cumulative reward (higher is better).}
\label{fig:real-world}
\end{figure}

\textbf{Experimental Results.} Based on Figure~\ref{fig:real-world}, DAL consistently performs well across real-world benchmarks, in both dynamic regret and cumulative reward with both GLR and GSR, giving the lowest or closely overlapping leading curves. We attribute this strong performance to the robustness DAL demonstrates in the synthetic settings, which capture a range of challenging NS scenarios. These findings underscore DAL's practical effectiveness. In what follows, we provide theoretical insights into its performance.

\section{Bandit-Driven Forced Exploration}
\label{sec:fx}
The previous sections examined DAL's stationary learner and detector. We now turn to its third component, forced exploration. Algorithm~\ref{alg:DAL} cycles through $\mathcal A_{\mathrm e}$ in a fixed order. This ensures that every probe is observed, but does not account for the evidence collected so far: after a change, some probes may reveal a much larger mean shift than others. We therefore ask whether choosing where to probe can improve detection and regret without changing the stationary learner.

\textbf{The Probe Selector.} We use a rule inspired by the controlled-sensing method of~\cite{huang2026ucbsensing}, which treats observation streams as arms and change evidence as a reward. Their known-model sensing rule has a first-order asymptotic guarantee; its unknown-mean GLR version is studied empirically. We investigate the latter idea inside DAL. Let $G_j$ and $N_j$ be the GLR statistic and observation count of probe $j$ since the last restart. Group the forced slots into blocks of $W=\lceil wN_{\mathrm e}\rceil$ slots, with $w\ge1$. At the start of every block, play every probe once in a fixed order. In each remaining slot, select the largest index
\begin{equation}
 \frac{G_j}{N_j}+s_j\sqrt{\frac{a\log u}{n_j}},
 \label{eq:fx_index}
\end{equation}
where $u$ is the slot's position in the block, $n_j$ is the number of observations of probe $j$ in that block, and $s_j$ is the root mean square of its GLR increments in the block. The first term favors evidence of a change; the bonus encourages sampling probes whose evidence is still variable. We set $G_j=0$ before two observations are available, and clear $G_j,N_j$ at every DAL restart.

\begin{algorithm}[t]
\small
\caption{Bandit-driven probe selection at a forced slot}\label{alg:fx}
\textbf{Input}: covering set $\mathcal A_{\mathrm e}$, block length $W=\lceil wN_{\mathrm e}\rceil$, bonus $a>0$\\
\textbf{Initialize once}: $u\leftarrow0$; $G_j,N_j,n_j,s_j\leftarrow0$ for all probes
\begin{algorithmic}[1]
\STATE $u\leftarrow u+1$
\STATE \textbf{if} {$u\le N_{\mathrm e}$} \textbf{then} select probe $j=u$
\STATE \textbf{else} select $j\in\arg\max_i G_i/N_i+s_i\sqrt{a\log u/n_i}$
\STATE Play $a^{(j)}$, update its statistics, and apply DAL's detector
\STATE \textbf{if} {DAL restarts} \textbf{then} $G_j,N_j\leftarrow0$ for all probes
\STATE \textbf{if} {$u=W$ or DAL restarts} \textbf{then} $u\leftarrow0$; $n_j\leftarrow0$, $s_j\leftarrow0$ for all probes
\end{algorithmic}
\end{algorithm}

Only the choice of action in a forced slot changes; the stationary learner, exploration frequency and restart rule remain the same. Every probe is still sampled in each block. This preserves coverage, but not necessarily the realized exploration regret, since the chosen actions and restart times can differ. The experiments below evaluate the direct selector in Algorithm~\ref{alg:fx}.

\textbf{Experimental Protocol.} We compare the two probe rules in DAL (GLR) on the PS-LBs of Section~\ref{sec:synth_exps}, with $\xi=0.8$, horizons $T\in\{5\times10^4,2\times10^5,5\times10^5,10^6\}$. Detectors use only probe observations. We use $15$ evaluation runs for $T\le2\times10^5$ and $10$ otherwise. Regarding the the bandit-based probe selector we set $a=0.01,w=1.5$.

\begin{figure*}[!htbp]
\centering
\includegraphics[width=0.8\linewidth]{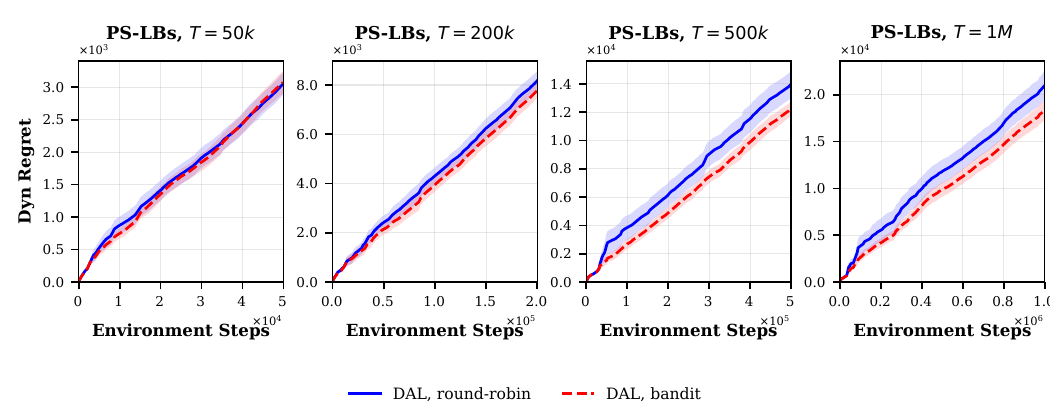}
\caption{Round-robin versus bandit-driven forced exploration in DAL (GLR), with $\xi=0.8$. Means and one standard error over $15$ runs for $T\le2\times10^5$ and $10$ otherwise.}
\label{fig:fx_curves}
\end{figure*}

\begin{figure*}[!htbp]
\centering
\begin{minipage}[t]{0.48\linewidth}
\centering
\includegraphics[width=0.75\linewidth]{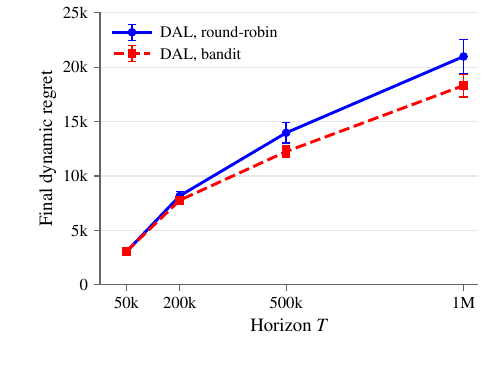}\\[-0.3em]
\small (a) Final dynamic regret
\end{minipage}\hspace{-0.1\linewidth}
\begin{minipage}[t]{0.48\linewidth}
\centering
\includegraphics[width=0.75\linewidth]{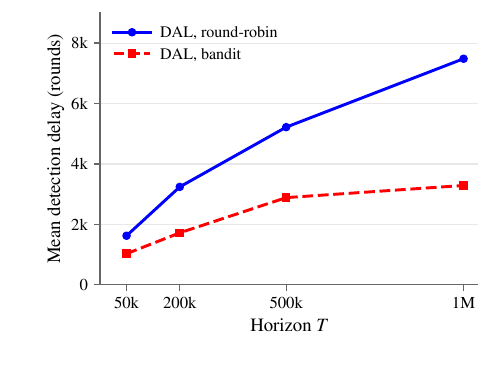}\\[-0.3em]
\small (b) Mean detection delay
\end{minipage}
\caption{Round-robin and bandit-driven forced exploration across horizons. Left: final regret with one-standard-error bars. Right: mean delay over detected changes.}
\label{fig:fx}\label{fig:fx_delay}
\end{figure*}

\textbf{Experimental Results.} At $T=50{,}000$, the main comparison gives nearly equal regret, $3055$ for round-robin and $3083$ for the bandit rule. At longer horizons, the benefit becomes clearer: regret decreases from $8176$ to $7775$ at $200{,}000$, from $13{,}975$ to $12{,}251$ at $500{,}000$, and from $20{,}989$ to $18{,}320$ at $10^6$, gains of approximately $5\%$, $12\%$ and $13\%$. Figures~\ref{fig:fx_curves} and~\ref{fig:fx} show the trajectories and horizon comparisons. With the smaller budget, the gains at the two longest horizons are approximately $16\%$ and $9\%$. Longer segments allow more repeated probing, making the allocation of observations more useful.

The detection ablation shows how probe allocation changes both the delay and the number of detected changes. Mean delay decreases from $1624$ to $1030$ rounds at $T=50{,}000$, while detected changes increase from $5.1$ to $6.3$ out of $9.3$. At $T=10^6$, delay decreases from $7482$ to $3288$, and detected changes increase from $13.1$ to $13.9$ out of $15.7$. At the intermediate horizons $T=200{,}000$ and $500{,}000$, the corresponding detected counts increase from $9.5$ to $10.1$ and from $11.2$ to $12.6$, respectively. Thus, the bandit rule detects approximately $6$--$24\%$ more changes at this exploration scale across the tested horizons.

\section{Summary and Outlook}
\label{sec:summary}
We introduced DAL, a practical, prior-free black-box framework for general NS bandits. Its plug-and-play design integrates seamlessly with a wide range of stationary bandit algorithms and different detectors, using forced exploration to monitor informative observations. The same design applies to parametric, kernelized, contextual and H\"older bandits, combinatorial bandits with semi-bandit or full-bandit feedback, and graph-feedback bandits.

On the theoretical side, we showed that DAL inherits and adapts the regret guarantees of its stationary input algorithm under detectable, sufficiently separated changes. A single regret bound retains the general stationary rate and makes the cost of exploration explicit. Appropriate covering sets yield the square-root rates in parametric and finite models, the benefits of item and graph feedback, and the instance-dependent rates of HBs. Regarding drifting non-stationarity, we hypothesized key conditions under which DAL excels---an insight further validated through experiments with different parametric reward models. 

Through extensive experiments in both PS and drifting settings, spanning synthetic and real-world benchmarks, DAL consistently performs well against both prior-free and prior-based baselines. Its leading performance in real-world scenarios highlights its value as a practical and effective solution. Finally, our study of forced exploration shows that learning where to probe can shorten detection delay and improve regret when detection depends on repeated probing.

While DAL advances both theory and practice, it opens new directions. First, a direct regret analysis in drifting environments would deepen our understanding of how accumulated variation affects detection and learning. Second, the current regret bounds rely on separation and detectable-shift conditions; weakening them would broaden the theoretical guarantees. The practical use of the learner's observations also motivates guarantees for adaptive monitoring without a separate prescheduled history. Deepening the study of piecewise stationarity may be the key to tackling these broader challenges and DAL can serve as a solid foundation towards that goal.

\bibliographystyle{IEEEtranN}
\bibliography{main}

\appendices

\section{Theoretical Results}
\label{app:theory}

\subsection{Proof of Proposition~\ref{prop:nspb_ae}}
\label{app:covers}
In NS-PBs and NS-CFBs, the mean reward at time $t$ is given by $f_t(a)=\mu(\langle\theta_t,a\rangle)$ for all $a\in\mathcal A$, where $\mu$ is injective; in CFBs, $\mu$ is the identity and actions are incidence vectors. Hence, equality of the mean rewards at an action is equivalent to equality of the corresponding inner products. This concerns the reward means, not inversion of noisy observations.

Suppose that $\mathcal A_{\mathrm e}\subseteq\mathcal A$ is a maximal linearly independent subset of $\mathcal A$. Every action is a linear combination of its elements. If the inner products of two parameters agree on $\mathcal A_{\mathrm e}$, they therefore agree on every action in $\mathcal A$, and injectivity of $\mu$ implies that the two reward functions coincide. Consequently, any change in the reward function changes the mean reward of at least one action in $\mathcal A_{\mathrm e}$. Thus, $\mathcal A_{\mathrm e}$ can be taken to be any maximal linearly independent subset of $\mathcal A$, with $N_{\mathrm e}=\dim\operatorname{span}(\mathcal A)\le d$; this dimension is denoted by $r$ in CFBs. The parameter itself need not be identifiable outside the action span.

\subsection{Proof of Proposition~\ref{prop:nskb_ae}}
In this subsection, we establish the construction of $\mathcal A_{\mathrm e}$ in the NS-KB setting. Recall that $\mathcal V_T$ consists of the centers of a $\delta_T$-cover of $\mathcal A$, with $\delta_T=L_T/(2BL_u)$. Let $[a]_{\mathrm e}$ denote the action in $\mathcal V_T$ closest to $a\in\mathcal A$. Then, we can leverage the assumed Lipschitz property of the permitted RKHS functions to obtain the following upper bound:
\begin{equation}
 |f(a)-f([a]_{\mathrm e})|\le BL_u\|a-[a]_{\mathrm e}\|_2\le BL_u\delta_T.
 \label{eq:Lip}
\end{equation}
For $\tilde a$ in the proposition and successive reward functions $f,f'$, the triangle inequality yields
\begin{align*}
 |f([\tilde a]_{\mathrm e})-f'([\tilde a]_{\mathrm e})|
 &\ge |f(\tilde a)-f'(\tilde a)|
 -|f(\tilde a)-f([\tilde a]_{\mathrm e})|-|f'(\tilde a)-f'([\tilde a]_{\mathrm e})|\\
 &\ge |f(\tilde a)-f'(\tilde a)|-2BL_u\delta_T
 =|f(\tilde a)-f'(\tilde a)|-L_T>0.
\end{align*}
This indicates that the value of the reward function at $[\tilde a]_{\mathrm e}$ must change by a nonzero amount. Thus, one can use observations from this action to detect the change. Taking the infimum over successive changes and using the strict inequality in the proposition gives a positive monitored margin. For the covering number on a general subset of $[0,R]^d$, partition the box into cells of side at most $\delta_T/\sqrt d$ and choose one action from each nonempty cell. There are at most $\lceil\sqrt dR/\delta_T\rceil^d$ cells. On the full box, cell centers with side length at most $2\delta_T/\sqrt d$ give the sharper bound. 

\subsection{Proof of Proposition~\ref{prop:nsHB_ae}}
Since $f-f'$ is continuous on $[0,1]^d$, choose $\tilde a$ with $|f(\tilde a)-f'(\tilde a)|=\|f-f'\|_\infty\ge L_T$. Let $[\tilde a]_{\mathrm e}$ be a nearest cover center. Each permitted reward function satisfies
$
 |f(\tilde a)-f([\tilde a]_{\mathrm e})|
 \le L\|\tilde a-[\tilde a]_{\mathrm e}\|_\infty^\beta
 \le L\delta_T^\beta.
$
Thus,
\[
 |f([\tilde a]_{\mathrm e})-f'([\tilde a]_{\mathrm e})|
 \ge |f(\tilde a)-f'(\tilde a)|-2L\delta_T^\beta
 \ge L_T-2L\delta_T^\beta>0.
\]
Taking the minimum over changes proves the margin bound. If $L_T\ge4L\delta_T^\beta$, the right-hand side is at least $L_T/2$. The witnessing point may differ between changes. The stated grid has radius $1/(2M)\le\delta_T$ in $\|\cdot\|_\infty$ and contains $(M+1)^d$ points. Its construction uses $\delta_T$, not $L_T$.

\subsection{Proof of Proposition~\ref{prop:struct_ae}}
For each item or arm $j$, fix a covering action that reveals its observation. Every cycle through $\mathcal A_{\mathrm e}$ therefore adds at least one observation to its history, whose mean is $\theta_{t,j}$. Hence the largest shift among the histories is $\|\theta-\theta'\|_\infty$. In CSBs,
$
 |f(a)-f'(a)|=\left|\sum_{j\in a}(\theta_j-\theta'_j)\right|
 \le K\|\theta-\theta'\|_\infty,
$
so $\|f-f'\|_\infty\ge L_T$ implies a monitored shift of at least $L_T/K$. Choosing one feasible revealing action for every active item, and removing duplicates, gives a CSB cover of at most $d$ actions. In top-$K$ selection, divide the items into groups of at most $K$ and complete the last group with already covered items if exactly $K$ items are required. This gives $\lceil d/K\rceil$ actions. In GFBs, $\|f-f'\|_\infty=\|\theta-\theta'\|_\infty$ and a maximal independent set is dominating: otherwise, an undominated vertex could be added. Its size is at most $\alpha(G)$, and $\alpha(G)\le\chi_{\mathrm{cl}}(G)$ since an independent set contains at most one vertex from each clique of a clique partition.

\subsection{Proof of Theorem~\ref{thrm:dal_reg}}
\label{app:thm_proof}
First, we partition the regret into two cases. If no false alarm occurs and all changes are detected within a short delay, we can separate the regret into three components: the regret due to forced exploration, the regret during the short detection (restart) delay after changes, and the regret incurred by the stationary bandit algorithm after the change is detected. If not, we use a crude linear bound to bound the regret and show that the probability of false alarm and that of late detection are low, which ensures that the regret due to detection failure is small~\citep{huang2025cdbppsmabs}.

We use the same decomposition for all the feedback models. For indexing the monitored histories within this proof, write
\[
 \mathcal J=\begin{cases}
 \mathcal C\times[N_{\mathrm e}],&\text{for scalar-reward histories},\\
 [d],&\text{for item or arm histories in CSBs and GFBs}.
 \end{cases}
\]
For $j=(c,i)$, let $\mathcal H_j=\mathcal H_{(c,a^{(i)})}$ and $\mu_{t,j}=f_t(c,a^{(i)})$; for an item or arm $j$, let $\mu_{t,j}=\theta_{t,j}$. The number $|\mathcal J|$ determines the false-alarm union bound, whereas $N_{\mathrm e}$ remains the number of forced actions per cycle. Fixed contexts correspond to $|\mathcal C|=1$.

It suffices to consider $|\mathcal C|N_{\mathrm e}(N_T+1)\le T$, since otherwise the claimed bound already exceeds $\bar\Delta T$. In this range clipping is inactive and the schedule satisfies $\alpha_k<1$ for $k\le N_T+1$. Consider a PS environment satisfying Assumption~\ref{assum:cp_assum}. Let $\tau_k$ be the $k$th detection point, with $\tau_0=0$:
\begin{equation}
 \tau_k\coloneqq\inf\left\{t>\tau_{k-1}:\mathcal D(\mathcal H_j)=\mathrm{detection}
 \text{ for some }j\in\mathcal J\right\}.
 \label{eq:detection_point}
\end{equation}
Recall that $\nu_0=1$ and $\nu_{N_T+1}=T+1$. Define
\begin{align}
 \mathcal G_k&\coloneqq
 \left\{\forall l\in[k-1],\ \tau_l\in\{\nu_l,\ldots,\nu_l+\ell_l-1\}\right\}
 \cap\{\tau_k\ge\nu_k\}, \label{eq:good_event_k}\\
 \mathcal E_k&\coloneqq
 \left\{\forall l\in[k-1],\ \tau_l\in\{\nu_l,\ldots,\nu_l+\ell_l-1\}\right\},
 \label{eq:good_event_E_k}
\end{align}
for $k\in[N_T+1]$, with $\mathcal E_1$ the universal space. Thus, $\mathcal G_k$ means that the previous changes were detected within their latency and there is no false alarm before $\nu_k$. A detection at $\nu_k$ is timely. First suppose every detection window ends before the horizon; the short final-segment case is handled below. Decomposing regret according to $\mathcal G_k$ gives
\begin{align}
 R_T
 &=\sum_{k=1}^{N_T+1}\mathbb E\!\left[
 \sum_{t=\nu_{k-1}}^{\nu_k-1}
 \big(\max_{\pi\in\Pi}f_t(C_t,\pi(C_t))-f_t(C_t,A_t)\big)\right]\nonumber\\
 &\le\sum_{k=1}^{N_T+1}\bar\Delta(\nu_k-\nu_{k-1})\mathbb P(\mathcal G_k^c)
 +\sum_{k=1}^{N_T+1}\mathbb E\!\left[\mathds1\{\mathcal G_k\}
 \sum_{t=\nu_{k-1}}^{\nu_k-1}
 \big(\max_{\pi\in\Pi}f_t(C_t,\pi(C_t))-f_t(C_t,A_t)\big)\right],
 \label{eq:reg_decomp}
\end{align}
where $\bar\Delta$ is the largest mean-reward gap over contexts, actions and time. Decomposing the bad event according to the first unsuccessful detection yields
\begin{align}
 \mathbb P(\mathcal G_k^c)
 &=\sum_{l=1}^{k-1}\mathbb P\big(\mathcal E_l,
 \tau_l\notin\{\nu_l,\ldots,\nu_l+\ell_l-1\}\big)
 +\mathbb P(\mathcal E_k,\tau_k<\nu_k)\nonumber\\
 &\le\sum_{l=1}^{k}\underbrace{\mathbb P(\tau_l<\nu_l\mid\mathcal E_l)}_{\Phi_1}
 +\sum_{l=1}^{k-1}\underbrace{\mathbb P(\tau_l\ge\nu_l+\ell_l\mid\mathcal E_l)}_{\Phi_2}.
 \label{eq:bad_union_thm1}
\end{align}
We bound $\Phi_1$ and $\Phi_2$ separately. In the conditional estimates below, first condition on the history through $\tau_{l-1}$, and then average over histories in $\mathcal E_l$. This fixes the phase of the forced schedule. Counts after that restart refer to its continuation without further resets, as in the definitions below.

\noindent$\bullet$ \emph{Upper-bounding $\Phi_1$.}
Recall that $\mathcal A_{\mathrm e}=\{a^{(i)}:i\in[N_{\mathrm e}]\}$. Continue the forced schedule after $\tau_{l-1}$ without further resets. Let $B_j(t)$ indicate whether its feedback at time $t$ is added to history $j$. For a scalar-reward history $j=(c,i)$,
\[
 B_{(c,i)}(t)=\mathds1\left\{C_t=c,\ (t-\tau_{l-1}-1)\bmod
 \left\lceil\frac{N_{\mathrm e}}{\alpha_l}\right\rceil=i-1\right\}.
\]
For an item or arm history, $B_j(t)=1$ when the prescribed forced action reveals that item or arm, and is zero on the learner's rounds. Define the $u$th observation time of history $j$ after $\tau_{l-1}$ by
\begin{equation}
 t'_{j,u}:=\inf\{t>t'_{j,u-1}:B_j(t)=1\},\qquad t'_{j,0}=\tau_{l-1},
 \label{eq:t'cis}
\end{equation}
and the number of observations added to it by time $t$ by
\begin{equation}
 n_j(t):=\sum_{v=\tau_{l-1}+1}^t B_j(v).
 \label{eq:pull}
\end{equation}
Let $\tau_j$ be its detector's stopping time in its own sample count,
\begin{equation}
 \tau_j:=\inf\{u\in\mathbb N:
 \mathcal D(\mathcal H_j)=\mathrm{detection}\text{ at time }t'_{j,u}\}.
 \label{eq:tau_ci}
\end{equation}
Each detector is continued even if another history triggers first; statistical independence between histories is not required. Before $\nu_l$, the monitored mean is $\mu_{\nu_{l-1},j}$. A union bound and the detector's false-alarm guarantee give
\begin{equation}
 \mathbb P(\tau_l<\nu_l\mid\mathcal E_l)
 \le\sum_{j\in\mathcal J}
 \mathbb P\big(\tau_j\le n_j(\nu_l-1)\mid\mathcal E_l\big)\le\sum_{j\in\mathcal J}
 \mathbb P_\infty(\tau_j\le T\mid\mathcal E_l)
 \le |\mathcal J|\delta_{\mathrm F}.
 \label{eq:false_alarm}
\end{equation}
The stationary continuation agrees with the actual law up to $\nu_l-1$. Conditional on the restart history and the future contexts, the observations of each fixed history are independent with a constant mean until that boundary. For item and arm histories, the assumption that the marginal law does not depend on which action reveals the outcome ensures the same property. Hence the false-alarm bound applies uniformly. Dependence between different items observed in one round does not affect this argument.

\noindent$\bullet$ \emph{Upper-bounding $\Phi_2$.}
Let $j^*$ be a history with the largest monitored mean shift at $\nu_l$:
\begin{equation}
 j^*\in\underset{j\in\mathcal J}{\arg\max}
 |\mu_{\nu_l,j}-\mu_{\nu_l-1,j}|.
\end{equation}
Its shift is at least $\Delta_{\mathrm c}$. Define the sufficient-sample events
\begin{align}
 \mathcal M_l\coloneqq
 \{n_{j^*}(\nu_l-1)-n_{j^*}(\tau_{l-1})\ge m_{\mathcal D}\},\:
 \mathcal L_l\coloneqq
 \{n_{j^*}(\nu_l+\ell_l-1)-n_{j^*}(\nu_l-1)\ge\ell_{\mathcal D}\}.
\end{align}
A late detection requires either insufficient observations or failure of the detector despite enough observations. Consequently,
\begin{equation}
 \label{eq:ld_bound}
 \mathbb P(\tau_l\ge\nu_l+\ell_l\mid\mathcal E_l)
 \le\mathbb P(\mathcal M_l^c\mid\mathcal E_l)
 +\mathbb P(\mathcal L_l^c\mid\mathcal E_l)
 +\mathbb P(\tau_l\ge\nu_l+\ell_l\mid
 \mathcal E_l\cap\mathcal M_l\cap\mathcal L_l).
\end{equation}
For each covering action $a^{(i)}$, every interval after the restart contains the following number of forced opportunities:
\begin{equation}
 \sum_{t=v+1}^{u}\mathds1\left\{(t-\tau_{l-1}-1)\bmod
 \left\lceil\frac{N_{\mathrm e}}{\alpha_l}\right\rceil=i-1\right\}
 \ge\left\lfloor\frac{u-v}{\lceil N_{\mathrm e}/\alpha_l\rceil}\right\rfloor.
 \label{eq:explore}
\end{equation}
Equality holds when $u-v$ is divisible by $\lceil N_{\mathrm e}/\alpha_l\rceil$. For each item or arm history, fix one covering action that reveals it, as in Proposition~\ref{prop:struct_ae}. Its observations satisfy the same lower bound, and other covering actions can only add more samples. On $\mathcal E_l$, Assumption~\ref{assum:cp_assum} implies $\tau_{l-1}\le\nu_l-m_l-1$, including $l=1$ since $\tau_0=0$ and $\nu_1>m_1$. Thus the last $m_l$ pre-change rounds lie after the restart. In CBs, write $j^*=(c^*,i^*)$. Conditioning on the restart history gives
\begin{align}
 &\mathbb E\big[n_{j^*}(\nu_l-1)-n_{j^*}(\nu_l-m_l-1)
 \mid\mathcal E_l,\tau_{l-1}\big]\nonumber\\
 &\quad=\sum_{t=\nu_l-m_l}^{\nu_l-1}\mathcal P_t(c^*)
 \mathds1\left\{(t-\tau_{l-1}-1)\bmod\left\lceil\frac{N_{\mathrm e}}{\alpha_l}\right\rceil=i^*-1\right\}
 \ge\frac{s m_l}{\lceil N_{\mathrm e}/\alpha_l\rceil},
 \label{eq:enough_pre_change_1}\\
 &\mathbb E\big[n_{j^*}(\nu_l+\ell_l-1)-n_{j^*}(\nu_l-1)
 \mid\mathcal E_l,\tau_{l-1}\big]
 \ge\frac{s\ell_l}{\lceil N_{\mathrm e}/\alpha_l\rceil}.
 \label{eq:enough_post_change_1}
\end{align}
The two windows contain exactly $m_l/\lceil N_{\mathrm e}/\alpha_l\rceil$ and $\ell_l/\lceil N_{\mathrm e}/\alpha_l\rceil$ opportunities. Applying Hoeffding's inequality to the independent context indicators in these windows and then averaging over $\tau_{l-1}$ yields
\begin{align}
 \mathbb P(\mathcal M_l^c\mid\mathcal E_l)
 &\le\exp\left[-\frac{2\big(s m_l/\lceil N_{\mathrm e}/\alpha_l\rceil-m_{\mathcal D}\big)^2}
 {m_l/\lceil N_{\mathrm e}/\alpha_l\rceil}\right]\le T^{-1},
 \label{eq:M_l_bound}\\
 \mathbb P(\mathcal L_l^c\mid\mathcal E_l)
 &\le\exp\left[-\frac{2\big(s\ell_l/\lceil N_{\mathrm e}/\alpha_l\rceil-\ell_{\mathcal D}\big)^2}
 {\ell_l/\lceil N_{\mathrm e}/\alpha_l\rceil}\right]\le T^{-1}.
 \label{eq:L_l_bound}
\end{align}
The last inequalities are precisely the quadratic inequalities solved by the two ceilings in Definition~\ref{def:sep_scaling}. The pre-change bound uses only the last $m_l$ rounds, rather than replacing the denominator for the whole pre-change history by that of a shorter window. For fixed contexts, including item and arm histories, both sufficient-sample events hold deterministically and their failure probabilities are zero.

For the third term of~\eqref{eq:ld_bound}, first fix the restart history and context sequence. The change index of the witnessing history is then fixed. On $\mathcal M_l\cap\mathcal L_l$, it has at least $m_{\mathcal D}$ pre-change samples and $\ell_{\mathcal D}$ post-change samples. Therefore,
\begin{align}
 &\mathbb P(\tau_l\ge\nu_l+\ell_l\mid\mathcal E_l\cap\mathcal M_l\cap\mathcal L_l)\nonumber\\
 &\quad\le\mathbb P\big(\tau_{j^*}>n_{j^*}(\nu_l+\ell_l-1)
 \mid\mathcal E_l\cap\mathcal M_l\cap\mathcal L_l\big)\nonumber\\
 &\quad\le\mathbb P\big(\tau_{j^*}>n_{j^*}(\nu_l-1)+\ell_{\mathcal D}
 \mid\mathcal E_l\cap\mathcal M_l\cap\mathcal L_l\big)
 \le\delta_{\mathrm D}.
 \label{eq:late_detect}
\end{align}
The final step is the latency guarantee, applied conditionally on the fixed sample times and then averaged. Substituting~\eqref{eq:M_l_bound}--\eqref{eq:late_detect} into~\eqref{eq:ld_bound} gives
\begin{equation}
 \mathbb P(\tau_l\ge\nu_l+\ell_l\mid\mathcal E_l)
 \le2T^{-1}+\delta_{\mathrm D}.
 \label{eq:late_total}
\end{equation}
Together with~\eqref{eq:false_alarm}, this gives
\begin{equation}
 \mathbb P(\mathcal G_k^c)\le|\mathcal J|k\delta_{\mathrm F}
 +(k-1)(2T^{-1}+\delta_{\mathrm D}).
 \label{eq:bad_event}
\end{equation}
This bounds the first term in~\eqref{eq:reg_decomp}. For the second term, define $\bar\alpha\coloneqq\max_{k\in[N_T+1]}\alpha_k$ and $\ell_0=0$. Separating the detection delay, forced exploration and stationary learner gives
\begin{align}
 &\mathbb E\!\left[\mathds1\{\mathcal G_k\}
 \sum_{t=\nu_{k-1}}^{\nu_k-1}
 \big(\max_{\pi\in\Pi}f_t(C_t,\pi(C_t))-f_t(C_t,A_t)\big)\right]\nonumber\\
 &\quad\le\bar\Delta\ell_{k-1}
 +\bar\Delta N_{\mathrm e}\left\lceil
 \frac{\nu_k-\nu_{k-1}}{\lceil N_{\mathrm e}/\alpha_k\rceil}\right\rceil\nonumber\\
 &\qquad+\mathbb E\!\left[\mathds1\{\mathcal G_k\}
 \sum_{\substack{t=\tau_{k-1}+1:\\(t-\tau_{k-1}-1)\bmod\lceil N_{\mathrm e}/\alpha_k\rceil\ge N_{\mathrm e}}}^{\nu_k-1}
 \big(\max_{\pi\in\Pi}f_t(C_t,\pi(C_t))-f_t(C_t,A_t)\big)\right]\nonumber\\
 &\quad\le\bar\Delta\ell_{k-1}
 +\bar\Delta\big[\bar\alpha(\nu_k-\nu_{k-1})+N_{\mathrm e}\big]
 +R_{\mathcal B}(\nu_k-\nu_{k-1}).
 \label{eq:bound_inter}
\end{align}
To justify the stationary term, condition on the forced observations, including their contexts, which determine the resets and $\mathcal G_k$. The analyzed DAL uses no learner observations for these decisions. With independent feedback across rounds and fresh learner initialization, its non-forced rounds after $\tau_{k-1}$ therefore form a fresh stationary run of $\mathcal B$ of length at most $\nu_k-\nu_{k-1}$. The expected bound and monotonicity of $R_{\mathcal B}$ apply to this run. The detector and learner observations are separate, as specified in Algorithm~\ref{alg:DAL}. Plugging~\eqref{eq:bad_event} and~\eqref{eq:bound_inter} into~\eqref{eq:reg_decomp} yields
\begin{align}
 R_T
 &\le\sum_{k=1}^{N_T+1}\bar\Delta(\nu_k-\nu_{k-1})
 \big[|\mathcal J|k\delta_{\mathrm F}+(k-1)(2T^{-1}+\delta_{\mathrm D})\big]\nonumber\\
 &\quad+\sum_{k=1}^{N_T+1}\left\{
 \bar\Delta\ell_{k-1}+\bar\Delta[\bar\alpha(\nu_k-\nu_{k-1})+N_{\mathrm e}]
 +R_{\mathcal B}(\nu_k-\nu_{k-1})\right\}\nonumber\\
 &\le\bar\Delta T|\mathcal J|(N_T+1)\delta_{\mathrm F}
 +2\bar\Delta N_T+\bar\Delta TN_T\delta_{\mathrm D}
 +\bar\Delta\sum_{k=1}^{N_T}\ell_k\nonumber\\
 &\quad+\bar\Delta[\bar\alpha T+(N_T+1)N_{\mathrm e}]
 +(N_T+1)R_{\mathcal B}\!\left(\frac{T}{N_T+1}\right).
 \label{eq:final_explicit}
\end{align}
The last step uses Jensen's inequality. If the last detection window extends to or past $T$, apply the same decomposition before the final change and charge the remaining rounds to $\bar\Delta\ell_{N_T}$. This gives the same bound without a terminal-segment separation condition.

Finally substitute the theorem's parameters. The exploration contribution satisfies
\[
 \bar\alpha T=\frac{\sqrt{|\mathcal C|N_{\mathrm e}(N_T+1)T}}{2\log^2T},
 \qquad (N_T+1)N_{\mathrm e}\le\sqrt{|\mathcal C|N_{\mathrm e}(N_T+1)T}.
\]
Moreover,
\[
 \left\lceil\frac{N_{\mathrm e}}{\alpha_k}\right\rceil
 \le 1+2\log^2T\sqrt{\frac{N_{\mathrm e}T}{k|\mathcal C|}},
 \qquad \sum_{k=1}^{N_T}k^{-1/2}\le2\sqrt{N_T}.
\]
In fixed-context settings this implies
\[
 \sum_{k=1}^{N_T}\ell_k\le\ell_{\mathcal D}
 \left[N_T+4\log^2T\sqrt{\frac{N_{\mathrm e}N_TT}{|\mathcal C|}}\right].
\]
For CBs, replace $\ell_{\mathcal D}$ by the ceiling factor in Definition~\ref{def:sep_scaling}, which is $\mathcal O((\ell_{\mathcal D}+\log T)/s^2)$. Property~\ref{proper:good_CD} gives $\ell_{\mathcal D}=\mathcal O(\log T)$. Thus the delay and exploration terms have the claimed order for fixed $s$. The failure terms are also dominated. For scalar-reward histories, $|\mathcal J|=|\mathcal C|N_{\mathrm e}$ and $\delta_{\mathrm F}=T^{-\eta}$. Using $|\mathcal C|N_{\mathrm e}(N_T+1)\le T$ and $\eta>1$ gives
\[
 T|\mathcal J|(N_T+1)\delta_{\mathrm F}
 \le\sqrt{|\mathcal C|N_{\mathrm e}(N_T+1)T}.
\]
For CSBs and GFBs, $|\mathcal J|=d$ and the chosen per-history level $\delta_{\mathrm F}=T^{-\eta}/d$ instead gives
\[
 T|\mathcal J|(N_T+1)\delta_{\mathrm F}
 =(N_T+1)T^{1-\eta}\le\sqrt{(N_T+1)T}.
\]
The sample requirement acquires an additional $\log d$ through $\log(1/\delta_{\mathrm F})$. Setting $\delta_{\mathrm D}=T^{-\eta}$ gives
\[
 TN_T\delta_{\mathrm D}+2N_T\le3\sqrt{N_TT}.
\]
Together with~\eqref{eq:final_explicit}, these inequalities complete the proof.

\subsection{Proof of Corollary~\ref{corr:dal_optimal}}
\label{sec:Cor_4.4_Proof}
By Theorem~\ref{thrm:dal_reg}, it suffices to bound the exploration term for each choice of $\mathcal A_{\mathrm e}$. The detector constants and context-recurrence parameter $s$ are fixed, as in the theorem.

In PS-PBs, the inputs in Table~\ref{table:reg_bounds} have $C_{\mathcal B}=\widetilde{\mathcal O}(d)$, and $N_{\mathrm e}\le d$ by Proposition~\ref{prop:nspb_ae}. Thus the $\sqrt{d(N_T+1)T}$ exploration term is absorbed by the stationary learning term. In PS-KBs, select a cover satisfying Proposition~\ref{prop:nskb_ae} with $N_{\mathrm e}=\mathcal O(C_{\mathcal B}^2)$. Then both terms are $\widetilde{\mathcal O}(C_{\mathcal B}\sqrt{(N_T+1)T})$. When a finite action set is smaller than the prescribed cover, it may instead be monitored in full.

In PS-CBs, $N_{\mathrm e}\le|\mathcal A|$. For the universal policy class, $|\Pi|=|\mathcal A|^{|\mathcal C|}$, so $|\mathcal C||\mathcal A|=\mathcal O(|\mathcal A|\log|\Pi|)$ when $|\mathcal A|\ge2$. This absorbs the exploration term into the finite-policy input's rate; the single-action case has zero regret. More generally, the covering condition in the corollary gives the same conclusion.

In PS-HBs, HOO with a compatible hierarchical partition has the stated stationary rate under the common, attained packing bound described in Section~\ref{sec:intro}~\citep[proof of Theorem~6]{bubeck2011xarmed}. The cover of Proposition~\ref{prop:nsHB_ae} has fixed size. Since $N_T+1\le T$ and $d_z\ge0$,
\[
 \sqrt{(N_T+1)T}\le
 (N_T+1)^{1/(d_z+2)}T^{(d_z+1)/(d_z+2)}.
\]
The exploration term is therefore dominated by the HOO input's restarted rate. Taking $d_z=d/\beta$ gives the worst-case H\"older exponent.

In PS-CSBs, $\bar\Delta\le K$ and $C_{\mathcal B}=\sqrt{dK}$ for the input in Table~\ref{table:reg_bounds}. Hence $N_{\mathrm e}=\mathcal O(d/K)$ gives $\bar\Delta\sqrt{N_{\mathrm e}}=\mathcal O(\sqrt{dK})$. For a general feasible family, the bound is instead
\[
 R_T=\widetilde{\mathcal O}\!\left((\sqrt{dK}+K\sqrt{N_{\mathrm e}})\sqrt{(N_T+1)T}\right).
\]

In PS-CFBs, $N_{\mathrm e}=r$ and $\bar\Delta\le K$. On the incidence span, $\|x_a\|_2\le\sqrt K$ and the parameter norm is at most $\sqrt d$. With ridge parameter $K$ and confidence-failure probability $T^{-2}$, the OFUL confidence and elliptical-potential bounds~\citep{abbasi-yadkori2011linucb} give
\[
 R_{\mathcal B}(n)=\widetilde{\mathcal O}\!\left(
 (\sqrt{Kd}+\sigma_\Sigma\sqrt r)\sqrt{rn}\right).
\]
Under the table's aggregate-noise condition this has $C_{\mathcal B}=\sqrt{Kdr}$, and $K\sqrt r\le\sqrt{Kdr}$ since $K\le d$. In PS-GFBs, a maximal independent set gives $N_{\mathrm e}\le\alpha(G)$ and UCB-N has $C_{\mathcal B}=\sqrt{\alpha(G)}$. These choices likewise absorb the exploration term, completing the proof.

\end{document}